%% file: main.tex
\pdfoutput=1

\documentclass[11pt]{article}

\usepackage[preprint]{acl}

\usepackage{times}
\usepackage{latexsym}

\usepackage[T1]{fontenc}

\usepackage[utf8]{inputenc}

\usepackage{microtype}

\usepackage{inconsolata}
\usepackage{amsmath}
\usepackage{booktabs}
\usepackage{multirow}
\usepackage{graphicx}
\usepackage{caption}
\usepackage{subcaption}
\usepackage{amsfonts}
\usepackage{algorithm}
\usepackage{algorithmic}
\usepackage{amssymb}
\usepackage{pifont}
\usepackage{xcolor}
\usepackage{tabularx}
\usepackage{bbding}
\usepackage{utfsym}
\usepackage{hyperref}
\usepackage{enumitem}
\usepackage{setspace}
\usepackage{courier} 
\usepackage{url}            
\usepackage{nicefrac}       
\usepackage{microtype}      
\usepackage[edges]{forest}

\definecolor{hidden-red}{RGB}{205, 44, 36}
\definecolor{hidden-blue}{RGB}{194,232,247}
\definecolor{hidden-orange}{RGB}{243,202,120}
\definecolor{hidden-green}{RGB}{34,139,34}
\definecolor{hidden-pink}{RGB}{255,245,247}
\definecolor{hidden-black}{RGB}{20,68,106}

\newcommand{\xmark}{\ding{55}}%
\newcommand{\eg}{E.g.,}
\newcommand{\xot}[0]{\textrm{XoT}}

\title{\textit{Navigate through Enigmatic Labyrinth}\\

A Survey of Chain of Thought Reasoning: Advances, Frontiers and Future}

\author{
    Zheng Chu$^{1}$\footnotemark[1], 
    Jingchang Chen$^{1}$\footnotemark[1], 
    Qianglong Chen$^{2}$\thanks{$\,$ Equal Contribution.},
    Weijiang Yu$^{2}$,
    Tao He$^{1}$\\
    \textbf{
    Haotian Wang$^{1}$, Weihua Peng$^{2}$, Ming Liu$^{1,3}$\thanks{$\,$ Corresponding Author.}, Bing Qin$^{1,3}$, Ting Liu$^{1}$
    } \\
    $^{1}$Harbin Institute of Technology, Harbin, China \\
    $^{2}$Huawei Inc., Shenzhen, China \\
    $^{3}$Peng Cheng Laboratory, Shenzhen, China \\
    \texttt{\{zchu,jcchen,mliu\}@ir.hit.edu.cn, chenqianglong.ai@gmail.com}
}

\begin{document}
\maketitle

\input{sections/abstract}
\input{sections/1.introduction}

\input{sections/2.background}
\input{sections/3.benchmarks}

\input{sections/4.methods}

\input{sections/5.frontiers}
\input{sections/6.future}

\input{sections/7.discussion}
\input{sections/8.conclusion}
\input{sections/limitation}
\input{sections/acknowledgement}

\bibliography{custom}

\input{sections/appendix}

\end{document}

%% file: sections/abstract.tex
\begin{abstract}\label{sec_abstract}
Reasoning, a fundamental cognitive process integral to human intelligence, has garnered substantial interest within 
artificial intelligence.
Notably, recent studies have revealed that chain-of-thought prompting significantly enhances LLM's reasoning capabilities, which attracts widespread attention from both academics and industry.
In this paper, we systematically investigate relevant research, summarizing advanced methods through a meticulous taxonomy that offers novel perspectives.
Moreover, we delve into the current frontiers and delineate the challenges and future directions, thereby shedding light on future research.
Furthermore, we engage in a discussion about open questions.
We hope this paper serves as an introduction for beginners and fosters future research.
Resources have been made publicly available at \href{https://github.com/zchuz/CoT-Reasoning-Survey}{https://github.com/zchuz/CoT-Reasoning-Survey}.

\end{abstract}

%% file: sections/1.introduction.tex
\section{Introduction} \label{sec:introduction}
In the realm of human cognition, reasoning stands as the linchpin, 
essential in the understanding of the world and the formation of our decisions.
As the scale of pre-training continues to expand~\citep{GPT3:TomBrown,GPT4:OpenAI,LLAMA:HuguoTouvron,LLAMA2:HugoTouvron}, large language models (LLMs) exhibit growing capabilities in numerous downstream tasks~\citep{Emergent:JasonWei, Emergent2:RylanSchaeffer, emergent-ability-survey}.
Recently, researchers have discovered that LLMs emerge with the capability for step-by-step reasoning through in-context learning, a phenomenon referred to as chain-of-thought (CoT) reasoning.
It is broadly observed that CoT prompting significantly boosts the reasoning abilities of LLMs, especially in complex tasks~\citep{FewshotCoT:JasonWei,Verifier:KarlCobbe,StrategyQA}.

Figure~\ref{figure:demo} illustrates an example of chain-of-thought reasoning. 
Rather than directly providing the answer, chain-of-thought reasoning offers a step-by-step reasoning trajectory.
Specifically, it decomposes intricate problems into manageable steps (\textit{thoughts}), simplifying the overall reasoning process, and creates a linkage (\textit{chain}) among the reasoning steps to ensure no important conditions are overlooked.
Additionally, chain-of-thought reasoning offers an observable reasoning process, allowing users to comprehend the model's decision-making trajectory and increase the trustworthiness and interpretability of the final answer.

\input{tables/fig_demo}

Benefiting from the remarkable performance of CoT prompting, it has attracted widespread attention across both academia and industry, evolving into a distinct research branch within the field of prompt engineering~\citep{PromptSurvey2:PengfeiLiu,PromptSurvey:ShuofeiQiao}. 
Moreover, it has emerged as a crucial component in the landscape of AI autonomous agents~\citep{AgentSurvey-1,AgentSurvey-2}.
However, these studies still lack a systematic review and analysis.
To fill this gap, we propose this work to conduct a comprehensive and detailed analysis of CoT reasoning.
Specifically, this paper delves into the broader scope of chain-of-thought reasoning, which we refer to as generalized chain-of-thought (XoT).
The core philosophy of XoT reasoning is the gradual unraveling of complex problems via a step-by-step reasoning approach.

Our contributions can be summarized as follows:
(1) \textbf{\textit{Comprehensive Survey}}: This is the first comprehensive survey dedicated for \xot{} reasoning;
(2) \textbf{\textit{Meticulous taxonomy}}: We introduce a meticulous taxonomy (shown in Figure~\ref{fig:taxonomy});
(3) \textbf{\textit{Frontier and Future}}: We discuss new frontiers, outline their challenges, and shed light on future research.
(4) \textbf{\textit{Resources}}: We make the resources publicly available to facilitate the research community.

\textbf{Survey Organization} We first give background and preliminary~(\S\ref{sec:background}); then present benchmarks~(\S\ref{sec_benchmarks}) and advanced methods~(\S\ref{sec_methods}) from different perspectives.
Furthermore, we discuss frontier research~(\S\ref{sec_frontiers}), and outline challenges as well as future directions~(\S\ref{sec_future}).
Finally, we give a further discussion about open questions~(\S\ref{sec:discussion}).

%% file: tables/fig_demo.tex
\begin{figure}[t]   
    \begin{center}
        \includegraphics[clip, width=\linewidth]{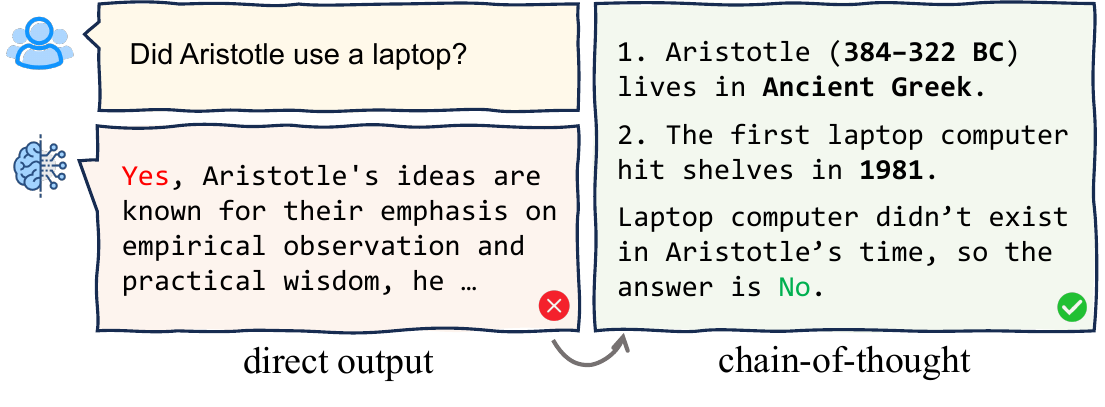}
        \caption{\label{figure:demo}
       The model tackles complex problems step-by-step under the guidance of chain-of-thought prompting.}
    \end{center}
\end{figure}

%% file: sections/2.background.tex
\section{Background and Preliminary} \label{sec:background}

\subsection{Background} \label{sec:background_background}
Over the past few years, 
as the scale of pre-training continuously increases~\citep{GPT3:TomBrown,BLOOM:TevenLeScao,LLAMA2:HugoTouvron,LLMSurvey:WayneXinZhao}, language models have emerged with numerous new capabilities, such as in-context learning~\citep{Emergent:JasonWei,GPT3:TomBrown} and chain-of-thought reasoning~\citep{FewshotCoT:JasonWei}. 
Accompanying this trend, pre-training then prompting has gradually replaced pre-training then fine-tuning as the new paradigm in natural language processing~\citep{LMSurvey:XipengQiu,LLMSurvey:WayneXinZhao}.

\subsection{Preliminary } \label{sec:background_preliminary}
In this section, we provide the preliminary for standard prompting and chain-of-thought reasoning.
Referring to \citet{PromptSurvey:ShuofeiQiao}, we define the notations as follows:
question $\mathcal{Q}$, prompt $\mathcal{T}$, probabilistic language model $p_{LM}$ and prediction $\mathcal{A}$.

First, we consider the few-shot standard prompting scenario, where prompt $\mathcal{T}_{SP}$ includes instruction $I$ and few-shot demonstrations (several question-answer pairs).
The model takes the question and prompt as inputs and produces the answer prediction $\mathcal{A}$ as its output, as shown in Equ.~(\ref{equ:icl_demo},\ref{equ:icl}).
\begin{align}
    &\label{equ:icl_demo}
    \mathcal{T}_{SP} = \{I,
        (x_1, y_1)
        , \cdots,
        (x_n, y_n)
    \} \\
    \label{equ:icl}
    &p(\mathcal{A} ~|~ \mathcal{T, Q}) =
    \prod_{i=1}^{|\mathcal{A}|}
    p_{LM}  
    (a_i ~|~ \mathcal{T, Q}, a_{<i})
\end{align}

Next, we consider chain-of-thought prompting under few-shot setting, wherein the prompt $\mathcal{T}_{CoT}$ includes instruction, questions, answers, and rationales $e_i$.
In chain-of-thought reasoning, the model no longer directly generates answers.
Instead, it generates step-by-step reasoning trajectories $\mathcal{R}$ before giving answers $\mathcal{A}$, as shown in Equ.~(\ref{equ:cot_demos},\ref{equ:cot},\ref{equ:cot_r},\ref{equ:cot_a}).
\begin{align}
    \label{equ:cot_demos}
    &\mathcal{T}_{\mathrm{CoT}} = \{I,
    (x_1, e_1, y_1),\cdots, (x_n, e_n, y_n)\} \\
    \noalign{\smallskip}
    \noalign{\smallskip}
    \label{equ:cot}
    &p(\mathcal{A, R} | \mathcal{T, Q}) =
    p(\mathcal{A} | \mathcal{T, Q, R})
    \cdot p(\mathcal{R} | \mathcal{T, Q}) \\
    &\label{equ:cot_r}
    p(\mathcal{R} ~|~ \mathcal{T, Q}) = 
    \prod_{i=1}^{|\mathcal{R}|}
    p_{LM} 
    (r_i ~|~ \mathcal{T, Q}, r_{<i}) \\
    &\label{equ:cot_a}
    p(\mathcal{A} | \mathcal{T,Q,R}) =
    \prod_{j=1}^{|\mathcal{A}|}
    p_{LM}  
    (a_i | \mathcal{T,Q,R}, a_{<j})
\end{align}

\subsection{Advantages of CoT Reasoning}
As a novel reasoning paradigm, chain-of-thought gains various advantages.
(1) Boosted Reasoning.
Chain-of-thought reasoning breaks down complex problems into manageable steps and establishes connections among these steps, thereby facilitating reasoning.
(2) Offering Interpretability.
Chain-of-thought reasoning provides observable reasoning traces, allowing the user to understand the model's decision, making the reasoning process transparent and trustworthy.
(3) Advance Collaboration.
Fine-grained reasoning traces facilitate user-system interaction, 
allowing for altering the model's execution trajectory, thereby fostering the development of autonomous agents powered by LLMs. 

%% file: sections/3.benchmarks.tex
\section{Benchmarks} \label{sec_benchmarks}

In this section, we briefly outline the benchmarks for evaluating reasoning capabilities, 
including mathematical, commonsense, symbolic, logical, and multi-modal reasoning.
The overview of benchmarks is shown in Table~\ref{tab:benchmarks}.
For more details about benchmarks, please refer to Appendix~\ref{sec:benchmarks}.

\input{tables/taxonomy_partial}

\paragraph{Mathematical Reasoning}
Mathematical reasoning forms the foundation of human intelligence, playing a crucial role in problem-solving, decision-making, and world comprehension.
It is commonly used to assess the general reasoning ability of LLMs~\citep{SVAMP:ArkilPatel,Verifier:KarlCobbe,MATH:DanHendrycks,LILA:SwaroopMishra}.
\paragraph{Commonsense Reasoning}
Commonsense reasoning is essential for the interaction in daily life and the perception of the world, which assesses the world comprehension capacity of language models~\citep{CommonsenseQA:AlonTalmor,CSQA2:AlonTalmor,StrategyQA}.
\paragraph{Symbolic Reasoning}
Symbolic reasoning disentangles semantics and serves as a testbed for language models' competence in simulating atomic operations~\citep{FewshotCoT:JasonWei,BigBench:AarohiSrivastava,BigBenchHard:MiracSuzgun}.
\paragraph{Logical Reasoning}
Logical reasoning is of paramount importance as it serves as the bedrock for rational thinking, robust problem-solving and interpretable decision-making~\citep{LogiQA:JianLiu,RClor:WeihaoYu,ProofWriter:OyvindTafjord,FOLIO:SimengHan}.
\paragraph{Multi-modal Reasoning}
Multimodal reasoning seamlessly integrates textual thought with sensory experiences from the natural world, such as visual scenes, and auditory sounds, to create a richer, more comprehensive understanding of information~\citep{VCR:RowanZellers,VisualCOMET,NEXT-QA:JunbinXiao,ScienceQA:PanLu,multimodal-cot-dataset-cure}.

%% file: tables/taxonomy_partial.tex
\tikzstyle{my-box}=[
    rectangle,
    draw=hidden-black,
    rounded corners,
    text opacity=1,
    minimum height=1.5em,
    minimum width=5em,
    inner sep=2pt,
    align=center,
    fill opacity=.5,
]
\tikzstyle{leaf}=[
    my-box, 
    minimum height=1.5em,
    fill=hidden-blue!90, 
    text=black,
    align=left,
    font=\normalsize,
    inner xsep=2pt,
    inner ysep=4pt,
]
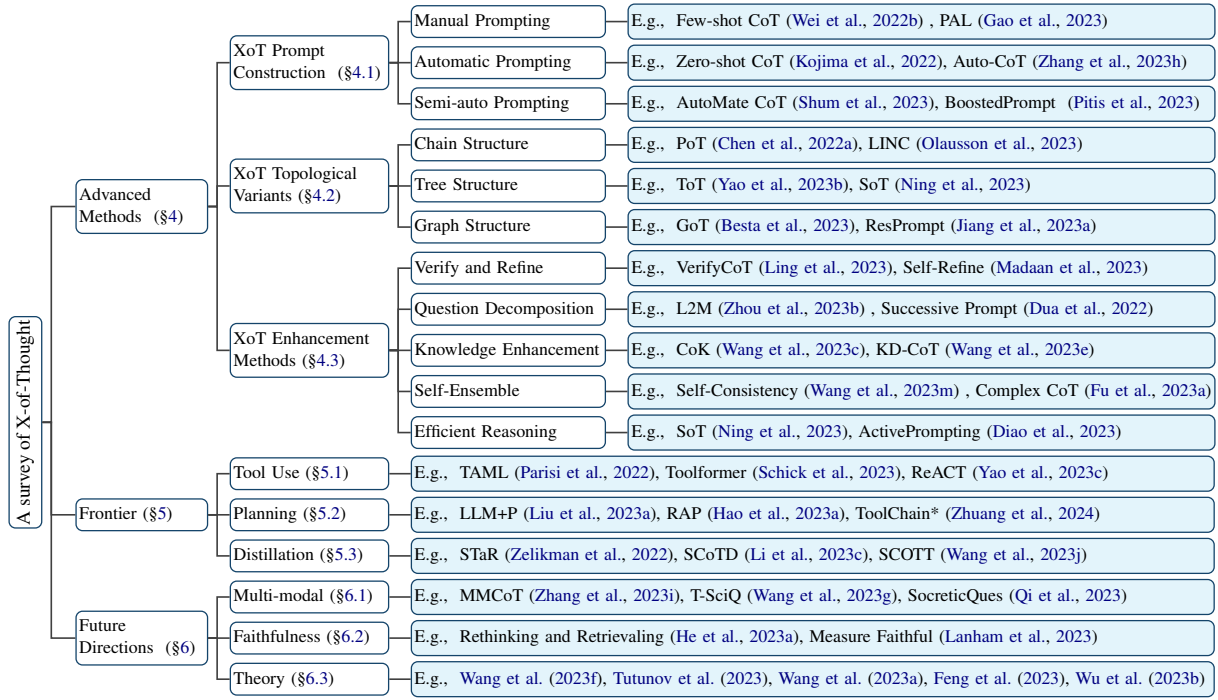
\begin{figure*}[t]
    \vspace{-2mm}
    \centering
    \resizebox{\textwidth}{!}{
        \begin{forest}
            forked edges,
            for tree={
                child anchor=west,
                parent anchor=east,
                grow'=east,
                anchor=west,
                base=left,
                font=\large,
                rectangle,
                draw=hidden-black,
                rounded corners,
                align=left,
                minimum width=4em,
                edge+={darkgray, line width=1pt},
                s sep=3pt,
                inner xsep=2pt,
                inner ysep=3pt,
                line width=0.8pt,
                ver/.style={rotate=90, child anchor=north, parent anchor=south, anchor=center},
            },
            where level=1{text width=7em,font=\normalsize,}{},
            where level=2{text width=8.5em,font=\normalsize,}{},
            where level=3{text width=10.5em,font=\normalsize,}{},
            where level=4{text width=12em,font=\normalsize,}{},
            [
                A survey of X-of-Thought, ver
                [
                    Advanced\\
                    Methods ~(\S\ref{sec_methods})
                    [
                        XoT Prompt \\Construction ~(\S\ref{sec_methods_construction})
                        [
                            Manual Prompting
                            [   
                                \eg~
                                Few-shot CoT~\cite{FewshotCoT:JasonWei} {,}
                                PAL~\cite{PAL:GapLuYu}
                                , leaf, text width=32.5em
                            ]
                        ]
                        [
                            Automatic Prompting
                            [
                                \eg~
                                Zero-shot CoT~\cite{ZeroshotCoT:TakeshiKojima}{,}
                                Auto-CoT~\cite{AutoCoT:ZhuoshengZhang}
                                , leaf, text width=32.5em
                            ]
                        ]
                        [
                            Semi-auto Prompting~
                            [
                                \eg~
                                AutoMate CoT~\cite{AutomateCoT:KashunShum}{,}
                                BoostedPrompt ~\cite{BoostedPromot:SilviuPitis}
                                , leaf, text width=32.5em
                            ]
                        ]
                    ]
                    [
                        XoT Topological \\Variants~(\S\ref{sec_methods_variants_structural})
                        [
                            Chain Structure
                            [
                            \eg~
                            PoT~\citep{PoT:WenhuChen}{,}
                            LINC~\citep{linc}
                            , leaf, text width=32.5em
                            ]
                        ]
                        [
                            Tree Structure
                            [
                            \eg~
                            ToT~\citep{ToT:ShunyuYao}{,}
                            SoT~\citep{SoT:XuefeiNing}
                            , leaf, text width=32.5em
                            ]
                        ]
                        [
                            Graph Structure
                            [
                            \eg~
                            GoT~\citep{GoTtext1:MaciejBesta}{,}
                            ResPrompt~\citep{Resprompt:ResidualConn}
                            , leaf, text width=32.5em
                            ]
                        ]
                    ]
                    [
                        XoT Enhancement \\Methods~(\S\ref{sec_methods_XoT_enhanced})
                        [
                            Verify and Refine
                            [
                                \eg~
                                VerifyCoT~\cite{VerifyCoT:ZhanLing}{,}
                                Self-Refine~\cite{SelfRefine:AmanMadaan}
                                , leaf, text width=32.5em
                            ]
                        ]
                        [
                            Question Decomposition
                            [
                                \eg~
                                L2M~\cite{Least-to-Most:DennyZhou}
                                {,}
                                Successive Prompt~\citep{SuccessivePrompt:DheeruDua}
                                , leaf, text width=32.5em
                            ]
                        ]
                        [
                            Knowledge Enhancement
                            [
                                \eg~
                                CoK~\cite{CoK2:JianingWang}{,}
                                KD-CoT~\cite{KD-CoT:KehengWang}
                                , leaf, text width=32.5em
                            ]
                        ]
                        [
                            Self-Ensemble
                            [
                                \eg~
                                Self-Consistency~\cite{SelfConsistency:XuezhiWang}
                                {,}
                                Complex CoT~\citep{ComplexCoT:YaoFu}
                                , leaf, text width=32.5em
                            ]
                        ]
                        [
                            Efficient Reasoning
                            [
                                \eg~
                                SoT~\cite{SoT:XuefeiNing}{,}
                                ActivePrompting~\cite{ActivePrompting:ShizheDiao}
                                , leaf, text width=32.5em
                            ]
                        ]
                    ]
                ]
                [
                    Frontier~(\S\ref{sec_frontiers})
                    [
                        Tool Use~(\S\ref{sec_frontiers_tool})
                        [
                            \eg~
                            TAML~\cite{TALM:Parisi}{,}
                            Toolformer~\cite{Toolformer:TimoSchick}{,}
                            ReACT~\cite{ReAct:Yao}
                            , leaf, text width=44.6em
                        ]
                    ]
                    [
                        Planning~(\S\ref{sec_frontiers_planning})
                        [
                            \eg~
                            LLM+P~\cite{LLMP:Liu}{,}
                            RAP~\citep{RAP:Hao2023}{,}
                            ToolChain*~\citep{toolchainstar}
                            , leaf, text width=44.6em
                        ]
                    ]
                    [
                        Distillation~(\S\ref{sec_frontiers_distill})
                        [
                            \eg~
                            STaR~\cite{STaR:Zelikman}{,}
                            SCoTD~\cite{SCoTD:Li}{,}
                            SCOTT~\cite{SCOTT:Wang}
                            , leaf, text width=44.6em
                        ]
                    ]
                ]
                [
                    Future \\ Directions ~(\S\ref{sec_future})
                    [
                        Multi-modal~(\S\ref{sec_future_multi_modal})
                        [
                            \eg~
                            MMCoT~\cite{MMCOT:ZhuoshengZhang}{,}
                            T-SciQ~\cite{T-SciQ:LeiWang}{,}
                            SocreticQues~\citep{SocraticQuestion}
                            , leaf, text width=44.6em
                        ]
                    ]
                    [
                        Faithfulness~(\S\ref{sec_future_faithfulness})
                        [
                            \eg~
                            Rethinking and Retrievaling~\cite{RethinkingRetrieval:HangfengHe}{,}
                            Measure Faithful~\cite{MeasureFaithful:TameraLanham}
                            , leaf, text width=44.6em
                        ]
                    ]
                    [
                        Theory~(\S\ref{sec_future_theory})
                        [
                            \eg~       
                            \citet{emnlp-best-paper}{,}
                            \citet{why-llm-correct-cot}{,}
                            \citet{CoTEmpiricalStudy:BoshiWang}{,}
                            \citet{TheoreticalPerspective:GuhaoFeng}{,}
                            \citet{AnalysisGradientCoT:SkylerWu}
                            , leaf, text width=44.6em
                        ]
                    ]
                ]
            ]
        \end{forest}
    }
    \caption{Taxonomy of Advanced Methods, Frontiers and Future Directions (Full version in Figure~\ref{fig:taxonomy_full}).}
    \label{fig:taxonomy}
\end{figure*}

%% file: sections/4.methods.tex
\section{Advanced Methods} \label{sec_methods}
This section discusses advanced XoT methods from three viewpoints: prompt construction~(\S\ref{sec_methods_construction}), 
topological variations~(\S\ref{sec_methods_variants_structural}), 
and enhancement methods~(\S\ref{sec_methods_XoT_enhanced}).
The taxonomy is shown in Figure~\ref{fig:taxonomy}.

\subsection{\xot{} Prompt Construction} \label{sec_methods_construction}
Based on the human effort for constructing chain-of-thought prompting,
we divide the construction approaches into three categories: 1) Manual \xot{}, 2) Automatic \xot{}, and 3) Semi-automatic \xot{}.

\subsubsection{Manual Prompting} \label{sec_methods_construction_manual}

\citet{FewshotCoT:JasonWei} first proposes chain-of-thought prompting (Fewshot CoT) by manually annotating natural language form rationales to guide models in stepwise reasoning. 
Moreover, \citet{ComplexCoT:YaoFu} discovers that using complex reasoning chains as demonstrations can further improve reasoning performance.
Yet, the NL form reasoning encounters inconsistent reasoning.
To mitigate intermediate errors in reasoning, PAL~\citep{PAL:GapLuYu}, PoT~\citep{PoT:WenhuChen}, MathPrompter~\citep{MathPrompter:ShimaImani} and NLEP~\citep{NaturalLanguageEmbedProgram} leverage rationales in programming language form, transforming problem-solving into program generation, and obtaining a deterministic answer through external program executor.
Although manual XoT demonstrates better performance, the annotation of rationales incurs a significant increase in cost and introduces dilemmas in demonstration selection.

\input{tables/fig_structure_variants}

\subsubsection{Automatic Prompting} \label{sec_methods_construction_auto}
Some work designs specific instructions to stimulate CoT reasoning under zero-shot, such as appending \textit{Let's think step by step} after questions~\citep{ZeroshotCoT:TakeshiKojima}.
There are also other types of instructions, including writing programs to solve problems~\citep{PoT:WenhuChen}, drafting plans before reasoning~\citep{PlanSolve:LeiWang}, generating meta instructions based on task information~\citep{AgentInstructs:Crispino} and role playing~\citep{naacl24-zeroshot-roleplay}.

However, due to the lack of guidance from clearly defined demonstrations, instruction-based methods appear extremely unstable.
Another route of work conducts few-shot reasoning based on automatically generated rationales (usually by zero-shot CoT), which improves the stability of reasoning.
These methods focus on selecting appropriate demonstrations.
\citet{AutoCoT:ZhuoshengZhang} chooses diverse rationales through clustering, 
\citet{MetaCoT} constructs demonstrations based on the question pattern, improving the generalization,
\citet{SelfAdapPrompting} employs answer entropy as a metric for selection, and \citet{Reprompting:WeijiaXu} uses Gibbs sampling to iteratively select demonstrations.

\subsubsection{Semi-automatic Prompting} \label{sec_methods_construction_semiauto}
Building upon automatic XoT based on few-shot learning, semi-automatic approaches incorporate a small number of human-annotated rationales to obtain supervised signals.
They focus on bootstrapping to acquire high-quality rationales and selecting appropriate demonstrations to facilitate reasoning.
\citet{SyntheticPrompting:ZhihongShao} generates high-quality rationales through alternating forward and backward synthetic processes, and \citet{BoostedPromot:SilviuPitis} iteratively expands the examples when encountering challenging questions, which mitigates the issue of limited human supervision. 
On the other hand, some studies optimize demonstration selection. \citet{AutomateCoT:KashunShum} and \citet{DynamicPrompting} utilize policy gradient optimization to learn demonstration selection strategy, 
while \citet{ExplanationSelection:XiYe} searches the development set and selects proper demonstration using two proxy metrics.

\subsubsection{Pros and Cons of Three Approaches}
Manual prompting relies on high-quality rationale annotations, which result in better performance. 
However, it encounters drawbacks such as high labor costs and challenges in domain transfer.
In contrast, automatic prompting incurs no labor costs and facilitates free domain transfer. 
However, it is plagued by errors and instability due to the absence of supervised signals.
Semi-automatic prompting strikes a dedicated balance, achieving a trade-off between performance and costs, making it more suitable for downstream applications.

\subsection{\xot{} Topological Variants} \label{sec_methods_variants_structural}

The evolution of \xot{} has led to the development of multiple topological variants\footnote{We consider \xot{} with chain structure and natural language rationales as vanilla CoT (the most primitive one).}.
In this section, we will delve into topological variants of \xot{}: chain structure, tree structure and graph structure.

\paragraph{Chain Structure}
The description format of rationales significantly influences reasoning execution. 
PAL~\citep{PAL:GapLuYu} and PoT~\citep{PoT:WenhuChen} use programming languages to depict the reasoning process, transforming problem-solving into code generation.
Similarly, formal logic description languages are also used to depict logical reasoning~\citep{linc,logic-lm,satlm}.
The aforementioned methods decouple the thought generation from execution, thereby eliminating inconsistency reasoning errors.
Additionally, algorithmic descriptions~\citep{AoT:BilgehanSel} can offer a high-level reasoning framework instead of details, endowing the model with the ability for global thinking.

\paragraph{Tree Structure}
Chain structure inherently limits the scope of exploration.
Through the incorporation of tree structures and search algorithms, models gain the capability to widely explore and backtrack during reasoning~\citep{ToT:JieyiLong, ToT:ShunyuYao}, as shown in Figure~\ref{figure:structure_variants}(e).
\citet{iclr24-tree-boost-of-thought} iteratively explores and evaluates multiple tree-of-thoughts to further enhance reasoning.
Benefiting from the exploration, tree variants have gained preliminary global planning capabilities towards the global optimum.
Meanwhile, \citet{Tree-of-Uncertain-T,prob-tot-mhqa} introduce uncertainty measurement based on Monte Carlo dropout and generation likelihood, respectively, thereby offering a more accurate evaluation of intermediate reasoning processes.
\citet{ThoughtPropagation} uses a bottom-up approach by building an analogy sub-problems tree.
In addition, \citet{SoT:XuefeiNing} initially delivers reasoning drafts, accelerating reasoning by solving tree structure sub-problems in parallel.
However, tree-based methods are restricted by demands of explicit question decomposition and state transition, which leads to limitations in task generalization.

\paragraph{Graph Structure}
Graph structures introduce loops and N-to-1 connections, enabling improved modeling of sub-problem aggregation and self-verification~\citep{GoTtext1:MaciejBesta,GoTtext2:BinLei}, as illustrated in Figure~\ref{figure:structure_variants}(f).
Graph structures outperform tree-based methods in handling complex problems. However, they rely on specially designed state decomposition, leading to poorer generalization.
To address this, 
\citet{Resprompt:ResidualConn} establishes an implicit graph upon the reasoning process through prompts, 
avoiding the constraints of explicit topological structures, thereby generalizing to various multi-step reasoning tasks.

The complex topological structure introduces a fine control flow, which facilitates LLMs in tackling harder problems.
However, this complexity also limits the application of these methods in general reasoning, posing a significant challenge that needs to be addressed in future research.

\subsection{\xot{} Enhancement Methods}
\label{sec_methods_XoT_enhanced}
This section introduces five enhanced XoT reasoning approaches, including verify and refine~(\S \ref{sec_methods_variants_refine}), question decomposition~(\S \ref{sec_methods_variants_decompose}), knowledge enhancement~(\S \ref{sec_methods_variants_knowledge}), self-ensemble~(\S \ref{sec_methods_variants_vote}) and efficient reasoning~(\S \ref{sec_methods_variants_efficiency}).

\subsubsection{Verify and Refine}  
\label{sec_methods_variants_refine}
\input{tables/fig_verify_refine}
LLMs tend to hallucinate, which manifests as factual and faithful errors in reasoning~\citep{hallucination-survey-huanglei}.
Incorporating verification and refinement can be an effective strategy for mitigating the phenomena.
In this section, we primarily focus on mitigating faithful errors, with a separate discussion of factual errors in the following knowledge enhancement section~(\S \ref{sec_methods_variants_knowledge}).

Reasoning can be refined based on critical feedback provided by LLMs.
\citet{REFINER:DebjitPaul} trains a small critic model to provide structured feedback, but the quality of the feedback is limited due to the model size.
\citet{SelfRefine:AmanMadaan} employs feedback from itself for iterative self-refinement, \citet{StepAwareVerifier} uses finer-grained feedback at the step level, and \citet{Reflexion:NoahShinn} further expands this method by incorporating long and short-term memory to provide more concise feedback.
However, recent research suggests that LLMs may not address issues beyond their own capabilities~\citep{knowtheyknow,knowdontknow}, which raises doubt on the effectiveness of self-feedback~\citep{CannotSelfCorrect}.
To remedy this, some work incorporates external feedback~\citep{critic:zhibin,maf} or performs secondary verification on the refined reasoning~\citep{SCREWS}.

On the other hand, logical reasoning structures are also well-suited for verification.
\citet{VerifyCoT:ZhanLing} devises a deductive reasoning form named Natural Program, which guarantees that the conclusion is derived from the designated premises.
\citet{coling24-refine-mitigating-misleading} applies a deductive filter to verify the entailment relationship between question and reasoning chains.
Some studies perform step-wise verification during the beam search decoding stage.
\citet{wirr-refine-self-guide-decoding} uses the log-probabilities of deductive reasoning as a search criterion, 
while \citet{wirr-refine-deductive-beamsearch} trains a deductive discriminator for verification.
Besides, backward (abductive) reasoning excels in detecting inconsistencies in reasoning.
It reconstructs conditions or variables in the question based on the reasoning chain to discover inconsistencies, thereby refining the reasoning~\citep{RCoT:TianciXue,SelfVerification:WengYixuan,FOBAR:WeisenJiang}.

Reasoning with LLMs is prone to hallucinations, and feedback from intermediate steps plays a crucial role in refining the reasoning. 
However, the current acquisition of feedback signals still has many shortcomings, which necessitates further research.

\subsubsection{Question Decomposition}  \label{sec_methods_variants_decompose}
\input{tables/fig_decomposition}
The philosophy of \xot{} is to solve questions step-by-step. However, vanilla CoT does not explicitly decompose questions, making it challenging to answer complex questions. 
To address this, certain approaches address intricate problems by progressively tackling straightforward sub-problems.

L2M~\citep{Least-to-Most:DennyZhou} initially breaks down the question into sub-questions in a top-down fashion. It then solves one sub-question at a time and leverages its solution to facilitate subsequent sub-questions.
\citet{SuccessivePrompt:DheeruDua} takes a similar approach to L2M, but it uses solutions from previous sub-questions to iteratively decompose questions.
\citet{DecomposedPrompt:TusharKhot} designs a modular task-sharing library that tailors more effective solutions to different classes of sub-questions.
\citet{coling24-decomposition-qmdr} breaks down the problem into a directed acyclic graph represented by QDMR, and then performs step-wise reasoning based on the graph dependencies.
In multi-hop reasoning, iterative decomposition has become a common practice~\citep{iCAP:BoshiWang,self-ask,IRCoT:Harsh}.
Additionally, some methods obtain a dedicated decomposer through supervised training rather than relying on the LLM itself~\citep{CoK1:XingxuanLi,cognitive-tree}.
However, when dealing with tabular reasoning, answering sub-questions may also pose a challenge, particularly when handling large tables. To tackle this issue, certain approaches involve decomposing both the questions and tables simultaneously~\citep{DecomposeTable,BindingLMSymbolic,naacl24-decomposition-tabsqlify}.

Bottom-up aggregation is also a viable solution, with a smaller exploration space.
\citet{SocraticQuestion} employs Socratic questioning for recursive self-questing to solve complex questions, while \citet{cumulative-reasoning}, in a similar fashion, breaks down the conditions of complex problems into small components and resolves them bottom-up.

It should be noted that both decomposition and aggregation are highly dependent on the proper problem division,
and reversely, a misaligned division
may yield counterproductive results.

\input{tables/fig_external_knowledge}
\subsubsection{Knowledge Enhancement} 
\label{sec_methods_variants_knowledge}

When dealing with knowledge-sensitive tasks, LLMs often make factual errors. 
Introducing external knowledge or mining the model's internal knowledge can help alleviate this issue.
Some methods explicitly utilize the model's intrinsic knowledge.
For example, \citet{Chain-of-Verification,MedVerify,StepBackPrompt} prompt models to output its parametric knowledge, and then reason based on it.
Additionally, \citet{iag} prompts the model to perform inductive reasoning on its internal knowledge, deriving more general conclusions.
Furthermore, \citet{Crystal} incorporates reinforcement learning to optimize introspective knowledge-grounded reasoning.
Meanwhile, \citet{MoT:XiaonanLi} leverages model's reasoning traces to construct a memory base, selecting relevant demonstrations whenever needed.

External knowledge is often more reliable than parametric knowledge. 
\citet{CoK1:XingxuanLi,KD-CoT:KehengWang} generates queries based on the question, utilizing a knowledge base as the external knowledge.
Building upon this, \citet{CoK2:JianingWang} introduces a verification step for the retrieved knowledge, further ensuring knowledge accuracy. 
However, when confronted with multi-hop reasoning, direct retrieval using the question can be insufficient. Therefore, \citet{self-ask,IRCoT:Harsh,iter-retgen,MCR:OriYoran} decompose the question and iteratively use sub-question for more precise retrieval.

\subsubsection{Self-Ensemble}  \label{sec_methods_variants_vote}

\input{tables/fig_vote_rank}

The sampling during generation introduces uncertainty, which in turn, creates the possibility of improving performance through self-ensemble.
\citet{Verifier:KarlCobbe} trains a verifier to rank answers, 
and \citet{coling24-selfensemble-rankprompt} utilizes LLMs to self-rank their predictions.
SC~\citep{SelfConsistency:XuezhiWang} performs majority voting based on answers across multiple samples, and \citet{ComplexCoT:YaoFu} proposes a complexity-based voting strategy on top of SC.
Widespread practical evidence indicates that self-ensemble is an effective way to improve performance. 
However, answer-based ensemble fails to consider intermediate steps. 
In response, \citet{SelfCheck:MiaoNing,MCR:OriYoran,Discriminator-Guided-GRACE} refines the ensemble at the step level,
and \citet{coling24-selfensemble-aggr-of-reasoning} introduces hierarchical answer aggregation.
Yet another concern is the limited diversity offered by probability sampling. To overcome this limitation, \citet{Diversity-of-Thought} uses different instructions, \citet{diverse-xot} ensembles various XoT variants, and \citet{cross-lingual-prompting} ensembles using multi-lingual reasoning chains.
Besides, the multi-agent debate (MAD) framework can also be regarded as heterogeneous ensemblings~\citep{mad1,mad2,mad3}.

Self-ensemble, as a simple yet effective means, has gained widespread favor.
Nevertheless, alongside the improvement in performance, there has been a multiplied increase in inference costs, which in turn limits its wide application.

\subsubsection{Efficient Reasoning}  \label{sec_methods_variants_efficiency}
LLMs are often inefficient in reasoning, such as high latency, substantial annotation costs, and elevated inference costs.
To speed up reasoning, \citet{SoT:XuefeiNing} decomposes the questions in parallel and handles them simultaneously, \citet{DraftVerify} generates a draft to skip intermediate layers during inference, and \citet{speculative_decoding,speculative_decoding_deepmind} introduce speculative decoding, which employs a smaller model for faster inference.
\citet{ActivePrompting:ShizheDiao} annotates high-uncertainty samples to reduce human costs, and \citet{AdaptiveConsistency:PranjalAggarwal} dynamically adjusts sampling frequency to reduce inference costs.
Further research should focus on efficient reasoning to promote the widespread application of LLMs.

%% file: tables/fig_structure_variants.tex
\begin{figure*}[t]   
    \begin{center}
        \includegraphics[clip, width=\linewidth]{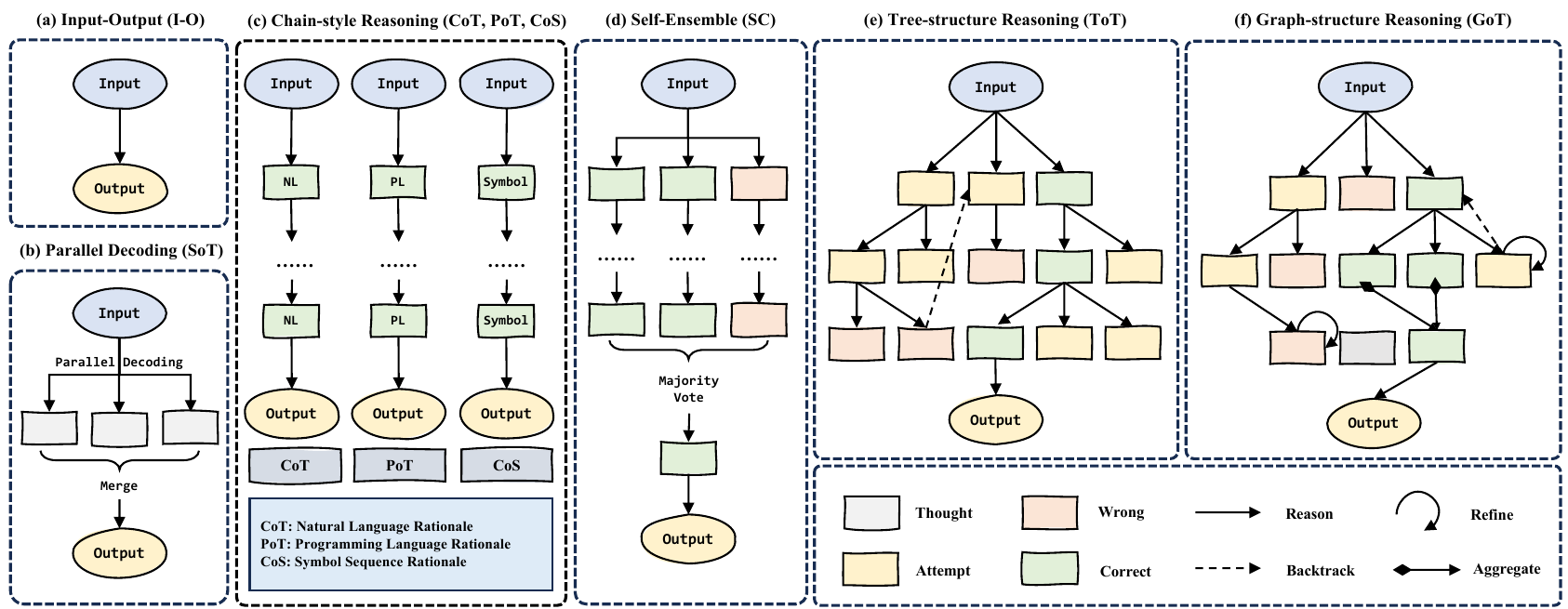}
        \caption{\label{figure:structure_variants}
        Topological variants emerging in the evolution of XoT. (a) standard I-O prompting, (b) parallel-constrained tree structure variants, (c) chain structure variants with distinct rationale descriptions, (d) chain structure variants with self-ensemble, (e) standard tree structure variants, and (f) standard graph structure variants.}
    \end{center}
\end{figure*}

%% file: tables/fig_verify_refine.tex
\begin{figure}[t]   
    \begin{center}
        \includegraphics[clip, width=\linewidth]{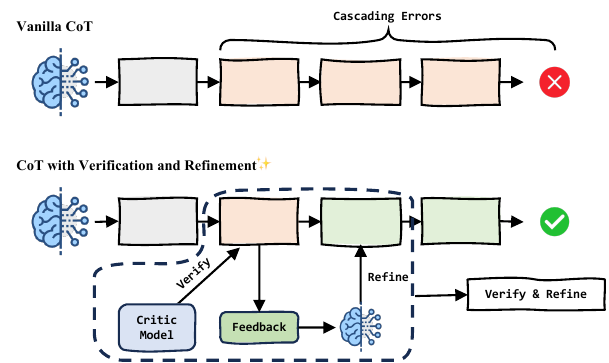}
        \caption{\label{figure:verify_refine}
        Verification and refinement rectify intermediate errors, which reduce cascading errors in reasoning.}
    \end{center}
\end{figure}

%% file: tables/fig_decomposition.tex
\begin{figure}[t]   
    \begin{center}
        \includegraphics[clip, width=\linewidth]{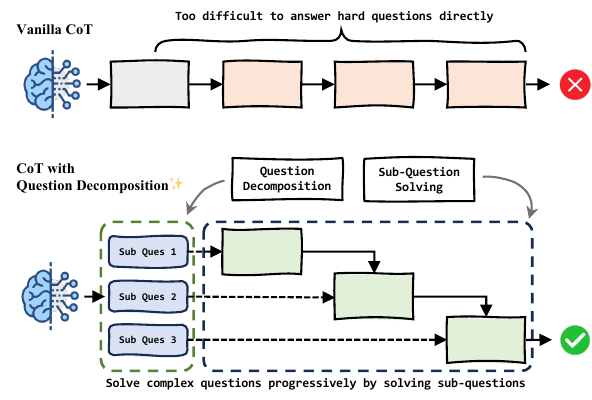}
        \caption{\label{figure:decompose}
       Question decomposition solves complex questions progressively by solving simple sub-questions.}
    \end{center}
\end{figure}

%% file: tables/fig_external_knowledge.tex
\begin{figure}[t]   
    \begin{center}
        \includegraphics[clip, width=\linewidth]{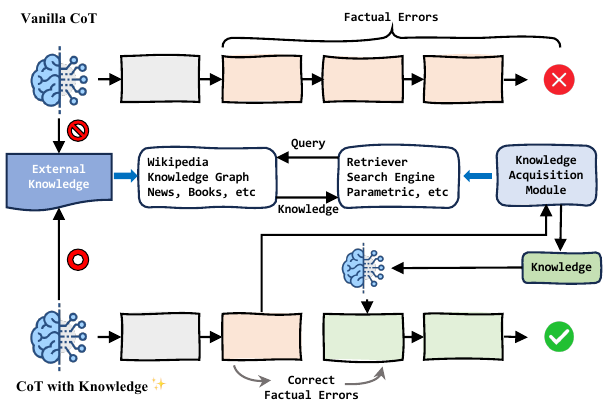}
        \caption{\label{figure:knowledge}
        Incorporating knowledge (either internal or external) helps mitigate factual errors in reasoning.
        }

    \end{center}
\end{figure}

%% file: tables/fig_vote_rank.tex
\begin{figure}[t]   
    \begin{center}
        \includegraphics[clip, width=\linewidth]{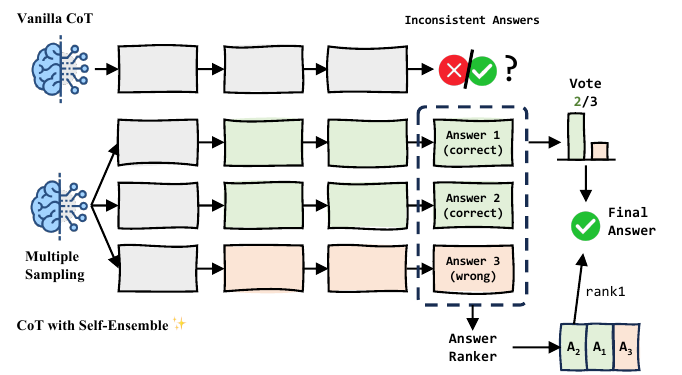}
        \caption{\label{figure:rank}
       Self-ensemble reduces inconsistency by selecting final answers from multiple samplings.}
    \end{center}
\end{figure}

%% file: sections/5.frontiers.tex
\section{Frontiers of Research} \label{sec_frontiers}

\subsection{Tool Use} \label{sec_frontiers_tool}

LLMs face difficulties in accessing news, performing calculations, and interacting with the environment.
Previous work endows LLMs with the ability to use external tools, 
enhancing their reasoning capabilities and enabling them to interact with the (multi-modal) external environment~\citep{TALM:Parisi,Toolformer:TimoSchick,HuggingGPT:Shen}.

However, these methods have limitations in facilitating multiple tool invocations and rectifying query errors.
To tackle this problem, ReAct~\citep{ReAct:Yao} and Reflexion~\citep{Reflexion:NoahShinn} integrate the strengths of reasoning and action to complement each other.
ART~\citep{ART} uses a task library to select relevant tools and reasoning demonstrations.
MM-REACT~\citep{MMREACT} further incorporates vision experts to facilitate multi-modal reasoning and action.

Above-mentioned studies focus on leveraging external tools to grant LLMs the capacities they initially lacked, 
thereby improving their performance across various domains.
Tool invocation facilitates interaction with external sources, enabling it to gather additional information, while
\xot{} enables effective elicitation, tracking, and action refining.

\subsection{Planning} \label{sec_frontiers_planning}

It is challenging for LLMs to provide accurate responses for complex goals,
which requires planning to decompose them into sub-tasks and track the execution process.
Plans can be described by code or definition languages.
\citet{AdaPlanner:Sun} generates Python code to control the agent, and iteratively refine the plan based on the execution feedback.
\citet{LLMP:Liu,LLMDP:Dagan} leverage the Planning Domain Definition Language (PDDL) ~\citep{PDDL:AlfonsoEmilio} to describe the planning procedure. 
PDDL assists in decomposing complex problems and utilizing specialized models for planning before converting the results into natural languages. 
\citet{ISR-LLM:Zhou} integrates self-refine~\citep{SelfRefine:AmanMadaan} with PDDL to achieve a better success rate in long-horizon sequential tasks.

Instead of pre-defined plans, many studies use search algorithms to dynamically plan and explore the action space.
Tree-of-Thought explores the problem through DFS or BFS search, and tracks and updates the intermediate states~\citep{ToT:ShunyuYao}.
RAP and LATS incorporate Monte Carlo Tree Search based on reasoning trajectories in planning~\citep{RAP:Hao2023,LATS:Zhou},
and ToolChain* enables more efficient exploring through heuristic A* search~\citep{toolchainstar}.

LLMs, endowed with robust reasoning capabilities, can devise strategies for achieving complex goals. 
Furthermore, the integration of planning, reasoning, memory, and tool utilization serves as a cornerstone for LLM-powered autonomous agents.

\subsection{Distillation of Reasoning Capabilities} \label{sec_frontiers_distill}

In low-resource scenarios such as edge computing, distillation offers a possibility for deploying LLMs. 
Some methods employ self-distillation for self-improvement without external supervision.
\citet{SelfImprove:JiaxinHuang} employs self-consistency to generate reasoning chains from unlabeled data, followed by fine-tuning, enhancing its generalized reasoning capabilities.
\citet{STaR:Zelikman} improves LM's reasoning capabilities via self-loop bootstrapping.

Despite the powerful reasoning exhibited by CoT, it emerges primarily in large-scale LLMs, with its usage limited in smaller models.
\citet{TeachingSmallLM:LucieCharlotte} finds that smaller models, after fine-tuning on CoT reasoning data, can also exhibit the capacity for step-by-step reasoning.
Following this trend, numerous studies attempt to distill the step-by-step reasoning capabilities of LLMs into smaller models.
\citet{DistillingStep:Hsieh} employs self-consistency to filter predictions, distilling high-quality reasoning chains from LLMs.
\citet{ReasoningTeacher,SCoTD:Li} find that sampling multiple reasoning chains per instance is paramount for improving students' reasoning capability.
SCOTT~\citep{SCOTT:Wang} utilizes contrastive decoding~\citep{ContrastiveDecoding:Xiang,ContrastiveDecoding:Sean} and counterfactual reasoning objective to tackle the shortcut problem.
\citet{coling24-distill-moe-distill} improves the generalization of reasoning for unseen tasks through LoRA mixture-of-experts distillation.

Recent studies have found that the reasoning capabilities of small models can be further improved by optimizing over preference data.
DialCoT~\citep{DialCoT:Han} decomposes reasoning steps into a multi-round dialog and optimizes the correct reasoning traces using PPO.
\citet{wirr-distill-preference-deepseek,coling24-distill-self-motivated-learning} train a reward model on automatically generated data, which is designed to rank LLM's reasoning traces,
and then optimizes smaller models using PPO.
\cite{wirr-distill-preference-mcts} utilizes Monte Carlo Tree Search to sample and score reasoning trajectories, generates preference data on the fly,
and uses DPO for online preference optimization.

Since code serves as an excellent intermediate representation for reasoning, \citet{naacl24-distill-program} distills program-aided reasoning capability into smaller models.
Meanwhile, some studies find that distilling reasoning chains from both natural language and code formats leads to further improvement~\citep{wirr-distill-mixdistill-1,wirr-distill-mixdistill-2}.
In addition to regular reasoning, 
\citet{coling24-distill-table-distillation} attempts to distill tabular reasoning capabilities,  
and \citet{coling24-distill-probe-retrieval} seeks to endow smaller models with retrieval-augmented reasoning capabilities.

These studies adopt a shared paradigm that distills smaller models with reasoning chains generated from larger models with superior reasoning capabilities. 
However, it is worth noting that language models have intricate tradeoffs associated with multi-dimensional capabilities, and distilling task-specific reasoning ability may adversely downgrade the general performance~\citep{SpecializingSL:Fu}.

%% file: sections/6.future.tex
\section{Future Directions} \label{sec_future}

Despite \xot{} reasoning has showcased remarkable performance on numerous tasks, there are still some challenges that necessitate further research.

\subsection{Multi-modal Reasoning} \label{sec_future_multi_modal}
Current XoT research mostly focuses on plain text. However, interacting with the real world necessitates multi-modal capabilities. 
To facilitate research, SciQA~\citep{ScienceQA:PanLu} and CURE~\citep{multimodal-cot-dataset-cure} are introduced to emphasize multi-modal CoT reasoning.
Through fine-tuning with the combination of vision and language features, \citet{MMCOT:ZhuoshengZhang,T-SciQ:LeiWang} endow models with multi-modal CoT reasoning capabilities, and \citet{GoT:YaoYao,HoT:FanglongYao} further incorporate graph structures to model multi-hop relationships.
Other approaches convert images to captions and use LLM for prompt-based reasoning~\citep{MMREACT,ddcot}.
However, the limited capabilities of vision-language models constrain their performance in multi-step reasoning~\citep{Flamingo:Jean,BLIP-2:JunnanLi, Kosmos-2:ZhilinPeng}.

Several critical challenges remain to be addressed in future research, which we summarize as follows: 
(1) {Vision-text interaction}: How can visual and textual features be effectively integrated, than solely depending on captions?
(2) {Harnessing VLLMs}: How can we better apply LLM-based reasoning techniques to the multi-modal domain?
(3) {Video Reasoning}: How to expand into video reasoning with complex temporal dependencies?

\subsection{Faithful Reasoning} \label{sec_future_faithfulness}
Extensive research indicates that LLMs often engage in unfaithful reasoning, such as factual errors and inconsistent reasoning.
To address factual errors, one common approach is retrieval augmentation~\citep{IRCoT:Harsh,VerifyEdit:RuochenZhao}, but it requires appropriate timing and retrieval accuracy.
Compared to factual errors, inconsistencies are more difficult to identify~\citep{wirr-faithful-causal-analysis}. 
Common detection methods include deductive logic~\citep{FOBAR:WeisenJiang,RCoT:TianciXue,VerifyCoT:ZhanLing}, post-processing~\citep{RethinkingRetrieval:HangfengHe,Chain-of-Natural-Language-Inference}, and critic-based approaches~\citep{SelfRefine:AmanMadaan,maf}.
Among them, Neural-symbolic reasoning~\citep{PoT:WenhuChen,linc} is a widely used approach for reducing inconsistencies, and question decomposition~\citep{DecomposeFaithful:AnshRadhakrishnan} has also demonstrated its effectiveness to some degree.
Furthermore, \citet{SnowBall,MeasureFaithful:TameraLanham} investigate the factors influencing faithfulness from an empirical perspective.

Faithful reasoning encounters two significant challenges:
(1) {Detection}: How can unfaithful reasoning be accurately identified?
(2) {Correction}: How can one obtain accurate feedback and make correct refinements based on that feedback?

\subsection{Theoretical Perspective} \label{sec_future_theory}
The mechanism behind the CoT and ICL has not been clearly explained so far.
Some studies empirically explore the roles of CoT and ICL in reasoning, offering practical insights~\citep{CoTEmpiricalStudy:BoshiWang,TextPattern,SemanticSymbolic:XiaojunTang}.
Another line of work explores from a theoretical perspective.
\citet{DissectingCoT:YingcongLi,TheoreticalPerspective:GuhaoFeng,ExoressivePowerofTRFMCoT,nips23-theory-whythinkstep} investigate why CoT enhances reasoning abilities, while \citet{AnalysisGradientCoT:SkylerWu,why-llm-correct-cot,interpret-multi-step-reason,emnlp-best-paper} examine the mechanisms from a feature-based standpoint (information flow, attention, variables, etc.).
Additionally, there have been preliminary explorations of the emergence mechanism~\citep{Emergent2:RylanSchaeffer,emergent-ability-survey}.

At present, the exploration of CoT theories is still limited to the surface level.
There are still open questions that require further in-depth investigation.
(1) How does the \textbf{emergence capability} arise?
(2) \textbf{In what way} does CoT enhance reasoning compared to standard few-shot prompting?

%% file: sections/7.discussion.tex
\section{Discussion} \label{sec_discussion}
We delve into open questions about chain-of-thought reasoning, with the details discussion in Appendix~\ref{sec:discussion}.
The discussion encompasses three topics:
(a) How does chain-of-thought reasoning ability emerge with large-scale pre-training?
(b) How to provide accurate feedback for a model's reasoning and decision-making. 
(c) The implications of chain-of-thought reasoning for LLM-powered autonomous agents and AGI.

%% file: sections/8.conclusion.tex
\section{Conclusion} \label{sec_conclusion}

In this paper, we conduct a systematic survey of existing research on generalized chain-of-thought reasoning, offering a comprehensive review of the field.
Specifically, we meticulously categorize advanced methods, delve into current frontier research, highlight existing challenges, identify potential future research directions, and discuss open questions.
This paper is the first systematic survey dedicated to CoT reasoning.
We hope that this survey will facilitate further research in this area.

%% file: sections/limitation.tex
\section*{Limitations}
This study provides the first comprehensive survey of generalized chain-of-thought (XoT) reasoning.
Related work, benchmarks details and further discussion can be found in Appendix~\ref{sec:appendix},\ref{sec:benchmarks}.

We have made our best effort, but there may still be some limitations.
On one hand, due to page limitations, we can only provide a brief summary of each method without exhaustive technical details.
On the other hand, 
we primarily collect studies from $^*$ACL, NeurIPS, ICLR, ICML, COLING and arXiv, and there is a chance that we may have missed some important work published in other venues.
In the benchmarks section, we primarily list widely used datasets, and more complete benchmarks can be found in \citet{survey-evaluate}.
As of now, there is no definitive conclusion on open questions.
We will stay abreast of discussions within the research community, updating opinions and supplementing overlooked work in the future.

%% file: sections/acknowledgement.tex
\section*{Acknowledgements}
The research in this article is supported by the National Key Research and Development Project (2021YFF0901602), the National Science Foundation of China (U22B2059, 62276083), and Shenzhen Foundational Research Funding (JCYJ20200109113441941), Major Key Project of PCL (PCL2021A06).
Ming Liu is the corresponding author.

%% file: sections/appendix.tex
\appendix
\section{Appendix}
\label{sec:appendix}

\subsection{Related Survey}
\label{sec:related_survey}
\citet{LLMSurvey:WayneXinZhao} primarily focuses on the development of contemporary LLMs, while \citet{LMSurvey:XipengQiu} surveys about early PLMs.
Some works discuss reasoning in specific domains, such as mathematical reasoning~\citep{MathSurvey:PanLu}, common-sense reasoning~\citep{CommonsenseQA:AlonTalmor}, and logical reasoning~\citep{survey-logical-reasoning}.
\citet{hallucination-survey-huanglei,survey:sirensong-hallucination} conducts an investigation into potential hallucination phenomena in LLM's reasoning. 
\citet{ICLSurvey:QingxiuDong} discusses in-context learning techniques in the era of LLMs, and \citet{NLRSurvey:FeiYu} conducts a macroscopic investigation into natural language reasoning.
\citet{PromptSurvey2:PengfeiLiu} discusses prompt tuning,  \citet{PromptSurvey:ShuofeiQiao,CoTPromptStrategySurvey,survey-towards-reasoning} focus on prompt engineering and reasoning strategies, 
and \citet{survey-igniting} highlights the development from chain-of-thought reasoning to autonomous agents.
This repository\footnote{\href{https://github.com/Timothyxxx/Chain-of-ThoughtsPapers}{Timothyxxx/Chain-of-ThoughtsPapers}} also collects  chain-of-thought reasoning papers.

Distinct from the above-mentioned surveys, this paper focuses on generalized chain-of-thought (XoT) reasoning in the era of LLMs.
This is the first systematic investigation into XoT reasoning, and we hope our work can serve as an overview to facilitate future research.

\subsection{Further Discussion}
\label{sec:discussion}
\paragraph{Open Question: Does CoT ability originate from code data pre-training?}
This is a pending question, initially summarized by \citet{yaofulog} and widely circulated in the research community.
In the early stages, LLMs like GPT3~\citep{GPT3:TomBrown} (davinci) and OPT~\citep{opt} usually do not possess CoT capabilities, and they do not use or only incorporate a small amount of code data (not specialized) during pre-training.
Recent models often incorporate specialized code data during pre-training, such as GPT-3.5, LLaMA2~\citep{LLAMA2:HugoTouvron} (with approximately 8\% of code data during pre-training) and they all possess strong CoT capabilities.
Additionally, \citet{PAL:GapLuYu,PoT:WenhuChen} have found that the use of programming language form rationales can significantly enhance the model's performance on complex reasoning tasks.
Various indications point towards the source of CoT abilities lying in code data during pre-training. 

Recently, \citet{iclr24-code-help-reasoning} investigates the impact of code data on LLMs at different training stages, reaching the first qualitative conclusion supported by quantitative experimental results.
They find that mixing code data during the pre-training stage enhances general reasoning abilities, 
while doing that in the instruction fine-tuning stage endows task-specific reasoning abilities.

\paragraph{Open Question: How to provide precise feedback on model's reasoning or decisions?}
When dealing with multi-step reasoning or decision-making tasks, errors often occur in intermediate steps, and if these errors are not corrected promptly, they may lead to cascading errors.
Currently, the primary methods for obtaining feedback include feedback from model itself~\citep{SelfRefine:AmanMadaan,Reflexion:NoahShinn}, feedback from other models~\citep{REFINER:DebjitPaul}, feedback from the external environment~\citep{maf,critic:zhibin}, and feedback based on reinforcement learning~\citep{process-outcome-feedback,letsverifystepbystep,letsrewardstepbystep}.
However, some studies have raised doubts about the ability of LLMs to provide self-feedback~\citep{CannotSelfCorrect,self-incorrect}.
Generally speaking, certain issues exist in current methods.
(1) How dependable is the feedback generated by the model itself?
(2) Is there a fundamental distinction between feedback from other language models and self-feedback? 
(3) Does the feedback quality still remain constrained by the model's capability boundaries?
(4) How is external feedback for various scenarios pre-defined, and how can this be expanded to different scenarios?

In summary, there is currently no fully satisfying feedback approach and more research attention is needed on how to accurately obtain feedback signals from the model's intermediate reasoning.

\paragraph{Discussion: Towards (early) AGI}
AGI has been the long-standing ultimate aspiration in the realm of artificial intelligence.
Recent research on LLM-powered autonomous agents 
has successfully demonstrated a preliminary implementation of nascent artificial general intelligence (AGI).

\textbf{Synergy between reasoning and interaction.}
Equipped with robust language comprehension capabilities, LLMs can interact with the external world through text-based interactions using plugins (tools, KB query, search engine, etc.)~\citep{Toolformer:TimoSchick,HuggingGPT:Shen,ToolLLM}.
Combining powerful reasoning capabilities, LLMs have made significant strides in various planning and decision-making tasks~\citep{Reflexion:NoahShinn,ToT:ShunyuYao,toolchainstar}, catalyzing research on LLM-based autonomous agents~\citep{AgentSurvey-1,AgentSurvey-2,survey-igniting}.

\textbf{LLM acts as the Brain (Controller).}
In contrast to traditional AI, which concentrates on specific tasks, AGI seeks the ability to understand general tasks~\citep{BERT:JacobDevlin,vit}, covering a widespread spectrum.
Within LLM-powered AI, the LLM typically serves as the brain (or central controller), handling reasoning, planning and decision-making, while delegating specific execution to dedicated modules (tools, weak AI, etc.)~\citep{HuggingGPT:Shen,autogpt}.
LLM-powered AI has already diverged significantly from weak AI and is progressing toward human cognition and thinking.

While some studies suggest that LLMs represent an early manifestation of AGI~\citep{gpt4-agi,compression-agi}, there are also scholars who contend that LLMs may not progress into AGI due to factors such as auto-regressive modeling and limited memory.
As of now, there is still intense debate on whether LLMs can evolve into AGI.
But regardless, LLM-powered AI has embarked on a distinctly different path from traditional AI, evolving towards a more generalized direction.

\subsection{Early Attempts and Efforts in Specific Domains}
In this section, we list the early attempts of XoT reasoning and efforts focused on specific domains.

Before the concept of CoT was introduced~\citep{FewshotCoT:JasonWei}, some efforts were made to enhance reasoning performance through the use of rationales~\citep{early-attemp-1,early-attemp-2,early-attemp-3,early-attemp-4}.
After that, certain work has empirically demonstrated the effectiveness of chain-of-thought prompting~\citep{CanLMLearnContext,UnreliabilityICL,AskAnything} and \citet{MultilingualCoT} explores multi-lingual CoT reasoning.
Other work focuses on specific domains, such as machine translation~\citep{HumanTranslation}, sentiment analysis~\citep{SentimentCoT}, 
sentence embeddings~\citep{CoTBertEmbeddings},
summarization~\citep{SummarizationCoT},
arithmetic~\citep{RecursionOfThought}, 
tabular reasoning~\citep{LLMTableReasoner,ZeroshotTableCoT},
and backdoor attack~\citep{iclr24-batchain-attack}, 
etc.
\citet{InferImplicitRelation,DatasetAutoAnswer} provide benchmarks and resources.
Besides, some research utilizes specific pre-training to enhance reasoning~\citep{SolvingQuantitativeReasoning,JiuZhangPLMForMath}.

\subsection{Empirical Results}
We statistic the performance of various \xot{} methods in mathematics, commonsense, and symbolic reasoning, as shown in Table~\ref{table:empirical_results}.
We primarily collect the performance of GPT series models and the results are mainly from corresponding papers (some results are used as baselines in other papers).
It is worth noting that due to variations in model checkpoints and experimental setups, even the methods with the same backbone LLM \underline{may not be fairly comparable}. 
Therefore, this table only provides a rough trend of performance.

\input{tables/tasks}
\section{Details of Benchmarks}
\label{sec:benchmarks}
\subsection{Mathematical Reasoning} \label{sec_benchmarks_math}
Mathematical reasoning is often used to measure the reasoning power of a model. 
Early benchmarks contain simple arithmetic operations~\cite{AddSub:MohanmmadJavad,SingleEQ:Koncel-Kedziorski,MultiArith:RoySubhro,MAWPS:RikKoncel}.
\citet{AQuA:WangLing} labels the reasoning process in natural language form, and \citet{MathQA:AidaAmini} builds on AQUA by labeling the reasoning process in program form.
Later benchmarks~\citep{ASDiv:MiaoShenYun,SVAMP:ArkilPatel,Verifier:KarlCobbe,PAL:GapLuYu} contain more complex and diverse questions.
\citep{TATQA:FengbinZhu,FinQA:ZhiyuChen,ConvFinQA:ZhiyuChen} require reasoning based on the table content.
There are also competition-level benchmarks~\cite{MATH:DanHendrycks,LILA:SwaroopMishra,NumGLUE:SwaroopMishra} and reading comprehension form benchmarks~\cite{DROP:DuaDheeru,TheoremQA:WenhuChen}.

\subsection{Commonsense Reasoning} \label{sec_benchmarks_commonsense}
Commonsense reasoning entails the process of drawing inferences, forming judgments, and gaining insights based on widely known and commonly accepted world knowledge.
Acquiring and understanding commonsense knowledge presents a significant challenge for models engaged in commonsense reasoning.
Various benchmarks have been put forward to address these challenges, including 
commonsense understanding~\citep{CommonsenseQA:AlonTalmor,CSQA2:AlonTalmor,ARC:SumithraBhakthavatsalam,OpenBookQA:MihaylovTodor,StrategyQA,CosmosQA:LifuHuang,PIQA:YonatanBisk},
event temporal commonsense reasoning~\citep{Event2Mind:HannahRashkin,McTaco:BenZhou}
, and commonsense verification~\citep{ComV:WangCunxiang}.

\subsection{Symbolic Reasoning} \label{sec_benchmarks_symbolic}
Symbolic reasoning here refers specifically to the simulation of some simple operations, which are simple for humans yet challenging for LLMs.
Last letter concatenation, coin flip, and reverse list~\citep{FewshotCoT:JasonWei} are the most commonly used symbolic reasoning tasks.
In addition, the collaborative benchmark BigBench~\citep{BigBench:AarohiSrivastava} and BigBench-Hard~\citep{BigBenchHard:MiracSuzgun} also contain several symbolic reasoning datasets, such as state tracking and object counting.

\subsection{Logical Reasoning} \label{sec_benchmarks_logical}
Logical reasoning encompasses deductive reasoning, inductive reasoning, and abductive reasoning.
Deductive reasoning derives conclusions from general premises~\citep{LogiQA:JianLiu,RClor:WeihaoYu,ProofWriter:OyvindTafjord,FOLIO:SimengHan,naacl24-benchmark-logic-self-verification}.
Inductive reasoning derives general conclusions from special cases~\cite{DEER:ZonglinYang}.
Abductive reasoning gives rational explanations for observed phenomena~\cite{PrOntoQA}.
 
\subsection{Multi-modal Reasoning} \label{sec_benchmarks_multi_modal}
In the real world, reasoning also involves information in modalities other than text, with visual modalities being the most prevalent.
To this end, many benchmarks for visual multi-modal reasoning are proposed~\citep{VCR:RowanZellers,VisualCOMET,PMR:QingxiuDong, ScienceQA:PanLu}, and among them, ScienceQA~\citep{ScienceQA:PanLu} annotates reasoning process and is the most commonly used visual multi-modal reasoning benchmark.
Video multi-modal reasoning~\citep{VLEP:JieLei,CLEVRER:KexinYi,STAR:BoWu,NEXT-QA:JunbinXiao,Causal-VidQA:JiantongLi,NewsKVQA:PranayGupta} is more challenging as it introduces additional temporal information compared to visual multi-modal reasoning.

\subsection{Comprehensive Benchmarks}
Apart from the aforementioned individual datasets, there are also some comprehensive evaluation benchmarks.
Some works aim to provide a holistic evaluation of the general reasoning capabilities~\citep{BigBench:AarohiSrivastava,BigBenchHard:MiracSuzgun,mmlu,ceval,holisticeval}.
In addition, there are also some multi-task benchmarks that focus on specific reasoning abilities, such as logical reasoning~\citep{logiglue,glore} and temporal reasoning~\citep{timebench,tram}.

\input{tables/empirical_results}

\input{tables/taxonomy_whole}

%% file: tables/tasks.tex
\begin{table*}[htb]
\centering
\tiny
\setlength{\tabcolsep}{3.5pt}
\resizebox{\textwidth}{!}{
\begin{tabular}{llccccc}
\toprule
 Task &  Dataset & Size & Input & Output & Rationale & Description\\
\midrule
 & AddSub \cite{AddSub:MohanmmadJavad} & 395 & Question & Number  & Equation & Simple arithmetic \\  
 & SingleEq \cite{SingleEQ:Koncel-Kedziorski} & 508 & Question & Number & Equation & Simple arithmetic \\
 & MultiArith \cite{MultiArith:RoySubhro} & 600 & Question  & Number & Equation & Simple arithmetic \\
 & MAWPS \cite{MAWPS:RikKoncel} & 3,320 & Question & Number & Equation & Simple arithmetic \\
 & AQUA-RAT \cite{AQuA:WangLing} & 100,000 & Question & Option & Natural Language & Math reasoning with NL rationale  \\
 & ASDiv \cite{ASDiv:MiaoShenYun} & 2,305 & Question & Number & Equation & Multi-step math reasoning \\
 & SVAMP \cite{SVAMP:ArkilPatel} & 1,000 & Question & Number & Equation & Multi-step math reasoning  \\
 & GSM8K \cite{Verifier:KarlCobbe} & 8,792 & Question &  Number & Natural Language & Multi-step math reasoning \\
 & GSM-Hard \cite{PAL:GapLuYu} & 936 & Question &  Number & Natural Language &  GSM8K with larger number\\
 & MathQA \cite{MathQA:AidaAmini}  & 37,297 & Question & Number & Operation & Annotated based on AQUA \\
 & DROP \cite{DROP:DuaDheeru} & 96,567 & Question+Passage & Number+Span & Equation & Reading comprehension form \\ 
 & TheoremQA \cite{TheoremQA:WenhuChen}  & 800 &  Question+Theorem & Number & \xmark & Answer based on theorems \\
 & TAT-QA \cite{TATQA:FengbinZhu}  & 16,552 & Question+Table+Text & Number+Span & Operation & Answer based on tables\\
 & FinQA \cite{FinQA:ZhiyuChen}  & 8,281 & Question+Table+Text & Number & Operation & Answer based on tables \\
 & ConvFinQA \cite{ConvFinQA:ZhiyuChen}  & 3,892 & Question+Table+Dialog & Number & Operation & Multi-turn dialogs \\
 & MATH \cite{MATH:DanHendrycks} & 12,500 & Question & Number & Natural Language & Challenging competition math problems \\
 
\multirow{-18}{*}{Mathematical } & NumGLUE \cite{NumGLUE:SwaroopMishra} & 101,835 & Question+Text & Number+Span & \xmark & Multi-task benchmark \\
\multirow{-18}{*}{Reasoning} & LILA \cite{LILA:SwaroopMishra} & 133,815 & Question+Text & Free-form  & Program & Multi-task benchmark \\

\midrule
 & ARC \cite{ARC:SumithraBhakthavatsalam} & 7,787 & Question & Option & \xmark & From science exam \\
 & OpenBookQA \cite{OpenBookQA:MihaylovTodor} & 5,957 & Question+Context & Option & \xmark & Open-book knowledges \\
 & PIQA \cite{PIQA:YonatanBisk} & 21,000 & Goal+Solution & Option  & \xmark & Physical commonsense knowledge \\
 
 & CommonsenseQA \cite{CommonsenseQA:AlonTalmor} & 12,247 & Question & Option  & \xmark & Derived from ConceptNet \\
 & CommonsenseQA 2.0 \cite{CSQA2:AlonTalmor} & 14,343 & Question & Yes/No  & \xmark & Gaming annotation with high quality \\
 & Event2Mind \cite{Event2Mind:HannahRashkin}  & 25,000 & Event & Intent+Reaction  & \xmark & Intension commonsense reasoning \\
 & McTaco \cite{McTaco:BenZhou} & 13,225 & Question & Option  & \xmark & Event temporal commonsense reasoning \\
 & CosmosQA \cite{CosmosQA:LifuHuang} & 35,588 & Question+Paragraph & Option & \xmark & Narrative commonsense reasoning \\
 & ComValidation \cite{ComV:WangCunxiang} & 11,997 & Statement & Option  & \xmark & Commonsense verification \\
\multirow{-10}{*}{Commonsense } & ComExplanation \cite{ComV:WangCunxiang} & 11,997 & Statement & Option/Free-form  & \xmark & Commonsense explanation \\
\multirow{-10}{*}{Reasoning} & StrategyQA \cite{StrategyQA} & 2,780 & Question & Yes/No & \xmark & Multi-hop commonsense reasoning  \\

\midrule
 & Last Letter Concat. \cite{FewshotCoT:JasonWei} & - & Words & Letters & \xmark & Rule-based \\
  & Coin Flip \cite{FewshotCoT:JasonWei} & - & Statement & Yes/No & \xmark & Rule-based \\
  & Reverse List \cite{FewshotCoT:JasonWei} & - & List & Reversed List & \xmark & Rule-based \\

\multirow{-4}{*}{Symbolic }   & BigBench \cite{BigBench:AarohiSrivastava} & - & - & - & \xmark & Contains multiple symbolic reasoning datasets \\
\multirow{-4}{*}{Reasoning}   & BigBench-Hard 
\cite{BigBenchHard:MiracSuzgun} & - & - & - & \xmark & Contains multiple symbolic reasoning datasets \\

\midrule
 & ReClor \cite{RClor:WeihaoYu} & 6,138 & Question+Context & Option & \xmark & Questions from GMAT and LSAT \\
 & LogiQA \cite{LogiQA:JianLiu} & 8,678 & Question+Paragraph & Option & \xmark & Questions from China Civil Service Exam \\
 & ProofWriter \cite{ProofWriter:OyvindTafjord} & 20,192 & Question+Rule & Answer+Proof & Entailment Tree & Reasoning process generation \\
 & FOLIO \cite{FOLIO:SimengHan} & 1,435 & Conclusion+Premise & Yes/No & \xmark & First-order logic \\
\multirow{-4}{*}{Logical} & DEER \cite{DEER:ZonglinYang} & 1,200 & Fact & Rule & \xmark & Inductive reasoning \\
\multirow{-4}{*}{Reasoning} & PrOntoQA \cite{PrOntoQA} & -& Question+Context & Yes/No+Proccess & First-Order Logic & Deductive reasoning \\

\midrule
 & VCR \cite{VCR:RowanZellers} & 264,720 & Question+Image & Option & Natural Language & Visual commonsense reasoning \\
 & VisualCOMET \cite{VisualCOMET} & 1,465,704 & Image+Event & Action+Intent & \xmark &  Visual commonsense reasoning \\
 & PMR~\cite{PMR:QingxiuDong} & 15,360 & Image+Background & Option & \xmark & Premise-based multi-modal reasoning \\
 & ScienceQA~\cite{ScienceQA:PanLu} & 21,208 & Q+Image+Context & Option & Natural Language & Multi-modal reasoning with NL rationales \\
 & VLEP \cite{VLEP:JieLei} & 28,726 & Premise+Video & Option & \xmark & Video event prediction \\
 & CLEVRER \cite{CLEVRER:KexinYi} & 305,280 & Question+Video & Option/Free-form & Program & Video temporal and causal reasoning \\ 
 & STAR \cite{STAR:BoWu} & 600,000 & Question+Video & Option & \xmark & Video situated reasoning \\ 
 & NEXT-QA \cite{NEXT-QA:JunbinXiao} & 47,692 & Question+Video & Option & \xmark & Video temporal,causal,commonsense reasoning \\ 
\multirow{-9}{*}{Multimodal }  & Causal-VidQA \cite{Causal-VidQA:JiantongLi} & 107,600 & Question+Video & Free-form & Natural Language & Video causal and commonsense reasoning \\ 
\multirow{-9}{*}{Reasoning}  & News-KVQA \cite{NewsKVQA:PranayGupta} & 1,041,352 & Q+V+KG & Option & \xmark & Video reasoning with external knowledge \\

\bottomrule
\end{tabular}
} 
\caption{An overview of benchmarks and tasks on reasoning. }
\label{tab:benchmarks}
\end{table*}

%% file: tables/empirical_results.tex
\begin{table*}[t]
\centering
\setlength{\tabcolsep}{4pt}
\resizebox{\textwidth}{!}{
\begin{tabular}{lcc|cccc|cc|cc}
\toprule
\multirow{2}{*}{\textbf{Method}} & \multirow{2}{*}{\textbf{Setting}} & \multirow{2}{*}{\textbf{Backbone}} & \multicolumn{4}{c|}{\textbf{Mathematical}} & \multicolumn{2}{c|}{\textbf{Commonsense}} & \multicolumn{2}{c}{\textbf{Symbolic}} \\
& & & GSM8K & SVAMP & Asdiv & AQuA & CSQA & StrategyQA & LastLetterConcat & CoinFlip \\
\midrule

I-O Prompting~\citep{GPT3:TomBrown} & fewshot & text-davinci-002 & 19.7 & 69.9 & 74 & 29.5 & 79.5 & 65.9 & 5.8 & 49.0 \\
Fewshot CoT~\citep{FewshotCoT:JasonWei} & fewshot & text-davinci-002 & 63.1 & 76.4 & 80.4 & 45.3 & 73.5 & 65.4 & 77.5 & 99.6 \\
PoT~\citep{PoT:WenhuChen} & fewshot & text-davinci-002 & 80 & 89.1 & - & 58.6 & - & - & - & - \\
Complex CoT~\citep{ComplexCoT:YaoFu} & fewshot & text-davinci-002 & 72.6 & - & - & - & - & 77 & - & - \\
Automate CoT~\citep{AutomateCoT:KashunShum} & fewshot & text-davinci-002 & 49.7 & 73.3 & 74.2 & 37.9 & 76.1 & 67.9 & 58.9 & -  \\
\midrule
Fewshot CoT~\citep{FewshotCoT:JasonWei} & fewshot & text-davinci-003 & 66.83 & 69.06 & - & 29.13 & - & - & - & - \\
PHP~\citep{ProgressiveHint:ChuanyangZheng} & fewshot & text-davinci-003 & 79 & 84.7 & - & 58.6 & - & - & - & - \\
Self-consistency~\citep{SelfConsistency:XuezhiWang} & fewshot & text-davinci-003 & 67.93 & 83.11 & - & 55.12 & - & - & - & - \\
Active Prompt~\citep{ActivePrompting:ShizheDiao} & fewshot & text-davinci-003 & 65.6 & 80.5 & 79.8 & 48 & 78.9 & 74.2 & 71.2 & - \\
Synthetic Prompt~\citep{SyntheticPrompting:ZhihongShao} & fewshot & text-davinci-003 & 73.9 & 81.8 & 80.7 & - & - & - & - & - \\
FOBAR~\citep{FOBAR:WeisenJiang} & fewshot & text-davinci-003 & 79.5 & 86 & - & 58.66 & - & - & - & - \\
Boosted Prompting~\citep{BoostedPromot:SilviuPitis} & fewshot & text-davinci-003 & 71.6 & - & - & 55.1 & - & - & - & - \\
\midrule
Fewshot CoT~\citep{FewshotCoT:JasonWei} & fewshot & code-davinci-002 & 60.1 & 75.8 & 80.1 & 39.8 & 79 & 73.4 & 70.4 & 99 \\
Self-Consistency~\citep{SelfConsistency:XuezhiWang} & fewshot & code-davinci-002 & 78 & 86.8 & 87.8 & 52 & 81.5 & 79.8 & 73.4 & 99.5 \\
PAL~\citep{PAL:GapLuYu} & fewshot & code-davinci-002 & 72 & 79.4 & 79.6 & - & - & - & - & - \\
Resprompt~\citep{Resprompt:ResidualConn} & fewshot & code-davinci-002 & 66.6 & - & - & 45.3 & - & - & - & - \\
DIVERSE~\citep{StepAwareVerifier} & fewshot & code-davinci-002 & 82.3 & 87 & 88.7 & - & 79.9 & 78.6 & - & - \\
Least-to-Most~\citep{Least-to-Most:DennyZhou} & fewshot & code-davinci-002 & 68.01 & - & - & - & - & - & 94 & - \\
Boosted Prompting~\citep{BoostedPromot:SilviuPitis} & fewshot & code-davinci-002 & 83.3 & 88.6 & - & 61.7 & - & - & - & - \\
\midrule
Fewshot CoT~\citep{FewshotCoT:JasonWei} & fewshot & gpt-3.5-turbo & 76.5 & 81.9 & - & 54.3 & 78 & 63.7 & 73.2 & 99 \\
Self-consistency~\citep{SelfConsistency:XuezhiWang} & fewshot & gpt-3.5-turbo & 81.9 & 86.4 & - & 62.6 & - & - & - & - \\
MetaCoT~\citep{MetaCoT} & fewshot & gpt-3.5-turbo & 75.1 & 88.6 & - & 54.7 & 72.4 & 64.5 & 77.2 & 100 \\
Verify CoT~\citep{VerifyCoT:ZhanLing} & fewshot & gpt-3.5-turbo & 86 & - & - & 69.5 & - & - & 92.6 & - \\
Active Prompting~\citep{ActivePrompting:ShizheDiao} & fewshot & gpt-3.5-turbo & 81.8 & 82.5 & 87.9 & 55.3 & - & - & - & - \\
RCoT~\citep{RCoT:TianciXue} & fewshot & gpt-3.5-turbo & 84.6 & 84.9 & 89.3 & 57.1 & - & - & - & - \\
FOBAR~\citep{FOBAR:WeisenJiang} & fewshot & gpt-3.5-turbo & 87.4 & 87.4 & - & 57.5 & - & - & - & - \\
Memory-of-Thought~\citep{MoT:XiaonanLi} & fewshot & gpt-3.5-turbo & - & - & - & 54.1 & - & - & - & - \\
Adaptive-consistency~\citep{AdaptiveConsistency:PranjalAggarwal} & fewshot & gpt-3.5-turbo & 82.7 & 85 & 83 & - & - & 67.9 & - & - \\
Boosted Prompting~\citep{BoostedPromot:SilviuPitis} & fewshot & gpt-3.5-turbo & 87.1 & - & - & 72.8 & - & - & - & - \\
\midrule

Zeroshot CoT~\citep{ZeroshotCoT:TakeshiKojima} & zeroshot & text-davinci-002 & 40.5 & 63.7 & - & 31.9 & 64 & 52.3 & 57.6 & 87.8 \\
PoT~\citep{PoT:WenhuChen} & zeroshot & text-davinci-002 & 57 & 70.8 & - & 43.9 & - & - & - & - \\
AutoCoT~\citep{AutoCoT:ZhuoshengZhang} & zeroshot & text-davinci-002 & 47.9 & 69.5 & - & 36.5 & 74.4 & 65.4 & 59.7 & 99.9 \\
COSP~\citep{AdaptiveConsistency:PranjalAggarwal} & zeroshot & code-davinci-001 & 8.7 & - & - &  & 55.4 & 52.8 & - & - \\
Plan-and-Solve~\citep{PlanSolve:LeiWang} & zeroshot & text-davinci-003 & 58.2 & 72 & - & 42.5 & 65.2 & 63.8 & 64.8 & 96.8 \\
Agent-Instruct~\citep{AgentInstructs:Crispino} & zeroshot & gpt-3.5-turbo & 73.4 & 80.8 & - & 57.9 & 74.1 & 69 & 99.8 & 95.2 \\
Self-Refine~\citep{SelfRefine:AmanMadaan} & zeroshot & gpt-3.5-turbo & 64.1 & - & - & - & - & - & - & - \\
RCoT~\citep{RCoT:TianciXue} & zeroshot & gpt-3.5-turbo & 82 & 79.6 & 86 & 55.5 & - & - & - & - \\
\bottomrule
\end{tabular}
}
\caption{The performance of various XoT methods in commonly used mathematical, commonsense and symbolic reasoning benchmarks. It is worth noting that, due to variations in the experimental setups of different methods, their performances are not directly comparable. The table is used to provide an overall empirical insight.
}
\label{table:empirical_results}
\end{table*}

%% file: tables/taxonomy_whole.tex
\clearpage

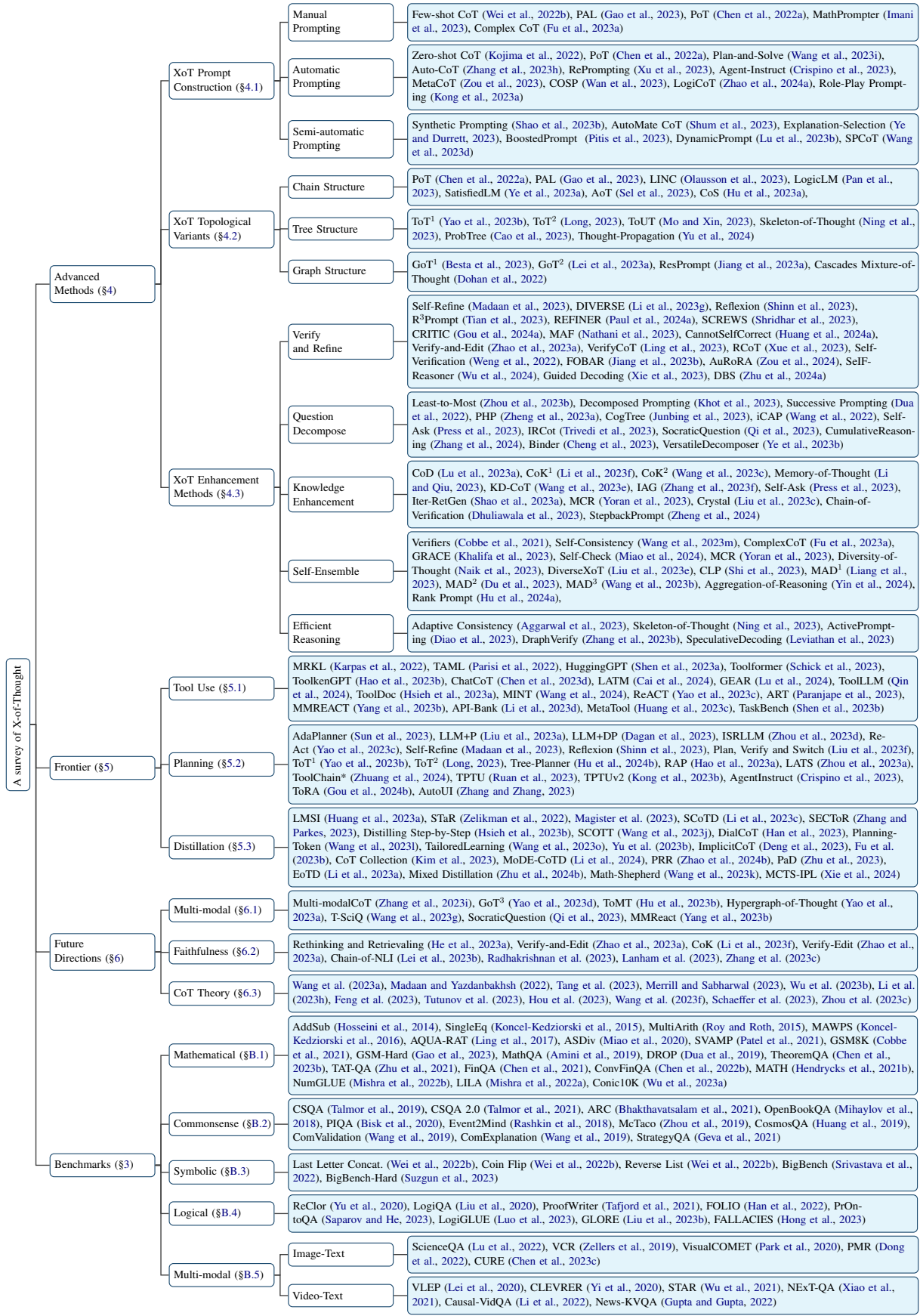
\begin{figure*}[t]
    \centering
    \resizebox{\textwidth}{!}{
        \begin{forest}
            forked edges,
            for tree={
                child anchor=west,
                parent anchor=east,
                grow'=east,
                anchor=west,
                base=left,
                font=\normalsize,
                rectangle,
                draw=hidden-black,
                rounded corners,
                minimum height=2em,
                minimum width=4em,
                edge+={darkgray, line width=1pt},
                s sep=3pt,
                inner xsep=0.4em,
                inner ysep=0.6em,
                line width=0.8pt,
                text width=8.5em,
                ver/.style={
                    rotate=90,
                    child anchor=north,
                    parent anchor=south,
                    anchor=center,
                    text width=11em
                },
                leaf/.style={
                    text opacity=1,
                    inner sep=2pt,
                    fill opacity=.5,
                    fill=hidden-blue!90, 
                    text=black,
                    text width=44.5em
                    font=\normalsize,
                    inner xsep=0.4em,
                    inner ysep=0.6em,
                    draw,
                }, 
            },
            [
                A survey of X-of-Thought, ver
                [
                    Advanced \\ Methods~(\S\ref{sec_methods})
                    [
                        XoT Prompt\\ Construction~(\S\ref{sec_methods_construction})
                        [
                            Manual \\Prompting
                            [   
                                Few-shot CoT~\cite{FewshotCoT:JasonWei}{,}
                                PAL~\cite{PAL:GapLuYu}{,}
                                PoT~\cite{PoT:WenhuChen}{,}
                                MathPrompter~\cite{MathPrompter:ShimaImani}{,}
                                Complex CoT~\cite{ComplexCoT:YaoFu}
                                , leaf, text width=44.5em
                            ]
                        ]
                        [
                            Automatic \\Prompting
                            [
                                Zero-shot CoT~\cite{ZeroshotCoT:TakeshiKojima}{,}
                                PoT~\cite{PoT:WenhuChen}{,}
                                Plan-and-Solve~\cite{PlanSolve:LeiWang}{,}
                                Auto-CoT~\cite{AutoCoT:ZhuoshengZhang}{,}
                                RePrompting~\cite{Reprompting:WeijiaXu}{,}
                                Agent-Instruct~\cite{AgentInstructs:Crispino}{,}
                                MetaCoT~\cite{MetaCoT}{,}
                                COSP~\cite{SelfAdapPrompting}{,}
                                LogiCoT~\cite{LogiCoT}{,}
                                Role-Play Prompting~\cite{naacl24-zeroshot-roleplay}
                                , leaf, text width=44.5em
                            ]
                        ]
                        [
                            Semi-automatic \\Prompting
                            [
                                Synthetic Prompting~\cite{SyntheticPrompting:ZhihongShao}{,}
                                AutoMate CoT~\cite{AutomateCoT:KashunShum}{,}
                                Explanation-Selection~\cite{ExplanationSelection:XiYe}{,}
                                BoostedPrompt ~\cite{BoostedPromot:SilviuPitis}{,}
                                DynamicPrompt~\citep{DynamicPrompting}{,}
                                SPCoT~\citep{spcot}
                                , leaf, text width=44.5em
                            ]
                        ]
                    ]
                    [
                        XoT Topological \\ Variants~(\S\ref{sec_methods_variants_structural})
                        [
                            Chain Structure
                            [
                                PoT~\citep{PoT:WenhuChen}{,}
                                PAL~\citep{PAL:GapLuYu}{,}
                                LINC~\citep{linc}{,}
                                LogicLM~\citep{logic-lm}{,}
                                SatisfiedLM~\citep{satlm}{,}
                                AoT~\citep{AoT:BilgehanSel}{,}
                                CoS~\citep{Chain-of-Symbol:HanxuHu}{,}
                                , leaf, text width=44.5em
                            ]
                        ]
                        [
                            Tree Structure
                            [
                                ToT${^1}$~\citep{ToT:ShunyuYao}{,}
                                ToT${^2}$~\citep{ToT:JieyiLong}{,}
                                ToUT~\citep{Tree-of-Uncertain-T}{,}
                                Skeleton-of-Thought~\citep{SoT:XuefeiNing}{,}
                                ProbTree~\citep{prob-tot-mhqa}{,}
                                Thought-Propagation~\citep{ThoughtPropagation}
                                , leaf, text width=44.5em
                            ]
                        ]
                        [
                            Graph Structure
                            [
                                GoT${^1}$~\citep{GoTtext1:MaciejBesta}{,}
                                GoT${^2}$~\citep{GoTtext2:BinLei}{,}
                                ResPrompt~\citep{Resprompt:ResidualConn}{,}
                                Cascades Mixture-of-Thought~\citep{LLMCascade}
                                , leaf, text width=44.5em
                            ]
                        ]
                    ]
                    [
                        XoT Enhancement \\ Methods~(\S\ref{sec_methods_XoT_enhanced})
                        [
                            Verify \\ and Refine
                            [
                                Self-Refine~\cite{SelfRefine:AmanMadaan}{,}
                                DIVERSE~\citep{StepAwareVerifier}{,}
                                Reflexion~\cite{Reflexion:NoahShinn}{,}
                                R$^3$Prompt~\citep{R3Prompt}{,}
                                REFINER~\cite{REFINER:DebjitPaul}{,}
                                SCREWS~\citep{SCREWS}{,}
                                CRITIC~\citep{critic:zhibin}{,}
                                MAF~\citep{maf}{,}
                                CannotSelfCorrect~\citep{CannotSelfCorrect}{,}
                                Verify-and-Edit~\cite{VerifyEdit:RuochenZhao}{,}
                                VerifyCoT~\cite{VerifyCoT:ZhanLing}{,}
                                RCoT~\citep{RCoT:TianciXue}{,}
                                Self-Verification~\citep{SelfVerification:WengYixuan}{,}
                                FOBAR~\citep{FOBAR:WeisenJiang}{,}
                                AuRoRA~\citep{coling24-refine-aurora}{,}
                                SeIF-Reasoner~\citep{coling24-refine-mitigating-misleading}{,}
                                Guided Decoding~\citep{wirr-refine-self-guide-decoding}{,}
                                DBS~\citep{wirr-refine-deductive-beamsearch}
                                , leaf, text width=44.5em
                            ]
                        ]
                        [
                            Question \\ Decompose
                            [
                                Least-to-Most~\cite{Least-to-Most:DennyZhou}{,}
                                Decomposed Prompting~\cite{DecomposedPrompt:TusharKhot}{,}
                                Successive Prompting~\cite{SuccessivePrompt:DheeruDua}{,}
                                PHP~\cite{ProgressiveHint:ChuanyangZheng}{,}
                                CogTree~\citep{cognitive-tree}{,}
                                iCAP~\cite{iCAP:BoshiWang}{,}
                                Self-Ask~\citep{self-ask}{,}
                                IRCot~\citep{IRCoT:Harsh}{,}
                                SocraticQuestion~\citep{SocraticQuestion}{,}
                                CumulativeReasoning~\citep{cumulative-reasoning}{,}
                                Binder~\citep{BindingLMSymbolic}{,}
                                VersatileDecomposer~\citep{DecomposeTable}
                                , leaf, text width=44.5em
                            ]
                        ]
                        [
                            Knowledge \\ Enhancement
                            [
                                CoD~\cite{CoD:HongyuanLu}{,}
                                CoK${^1}$~\cite{CoK1:XingxuanLi}{,}
                                CoK${^2}$~\cite{CoK2:JianingWang}{,}
                                Memory-of-Thought~\cite{MoT:XiaonanLi}{,}
                                KD-CoT~\cite{KD-CoT:KehengWang}{,}
                                IAG~\citep{iag}{,}
                                Self-Ask~\citep{self-ask}{,}
                                Iter-RetGen~\citep{iter-retgen}{,}
                                MCR~\citep{MCR:OriYoran}{,}
                                Crystal~\citep{Crystal}{,}
                                Chain-of-Verification~\citep{Chain-of-Verification}{,}
                                StepbackPrompt~\citep{StepBackPrompt}
                                , leaf, text width=44.5em
                            ]
                        ]
                        [
                            Self-Ensemble
                            [
                                Verifiers~\cite{Verifier:KarlCobbe}{,}
                                Self-Consistency~\cite{SelfConsistency:XuezhiWang}{,}
                                ComplexCoT~\cite{ComplexCoT:YaoFu}{,}
                                GRACE~\citep{Discriminator-Guided-GRACE}{,}
                                Self-Check~\cite{SelfCheck:MiaoNing}{,}
                                MCR~\cite{MCR:OriYoran}{,}
                                Diversity-of-Thought~\citep{Diversity-of-Thought}{,}
                                DiverseXoT~\citep{diverse-xot}{,}
                                CLP~\citep{MultilingualCoT}{,}
                                MAD${^1}$~\citep{mad1}{,}
                                MAD${^2}$~\citep{mad2}{,}
                                MAD${^3}$~\citep{mad3}{,}
                                Aggregation-of-Reasoning~\citep{coling24-selfensemble-aggr-of-reasoning}{,}
                                Rank Prompt~\citep{coling24-selfensemble-rankprompt}{,}
                                , leaf, text width=44.5em
                            ]
                        ]
                        [
                            Efficient \\ Reasoning
                            [
                                Adaptive Consistency~\cite{AdaptiveConsistency:PranjalAggarwal}{,}
                                Skeleton-of-Thought~\cite{SoT:XuefeiNing}{,}
                                ActivePrompting~\cite{ActivePrompting:ShizheDiao}{,}
                                DraphVerify~\citep{DraftVerify}{,}
                                SpeculativeDecoding~\citep{speculative_decoding}
                                , leaf, text width=44.5em
                            ]
                        ]
                    ]
                ]
                [
                    Frontier~(\S\ref{sec_frontiers})
                    [
                        Tool Use~(\S\ref{sec_frontiers_tool})
                        [
                            MRKL~\cite{MRKL:Karpas}{,}
                            TAML~\cite{TALM:Parisi}{,}
                            HuggingGPT~\cite{HuggingGPT:Shen}{,}
                            Toolformer~\cite{Toolformer:TimoSchick}{,}
                            ToolkenGPT~\cite{ToolkenGPT}{,}
                            ChatCoT~\cite{ChatCoT}{,}
                            LATM~\cite{LATM}{,}
                            GEAR~\cite{GEAR}{,}
                            ToolLLM~\cite{ToolLLM}{,}
                            ToolDoc~\cite{ToolDocumentation}{,}
                            MINT~\cite{MINT}{,}
                            ReACT~\cite{ReAct:Yao}{,}
                            ART~\cite{ART}{,}
                            MMREACT~\cite{MMREACT}{,}
                            API-Bank~\cite{APIBank:Li}{,}
                            MetaTool~\cite{MetaTool:Huang}{,}
                            TaskBench~\cite{TaskBench}
                            , leaf, text width=55.0em
                        ]
                    ]
                    [
                        Planning~(\S\ref{sec_frontiers_planning})
                        [
                            AdaPlanner~\cite{AdaPlanner:Sun}{,}
                            LLM+P~\cite{LLMP:Liu}{,}
                            LLM+DP~\cite{LLMDP:Dagan}{,}
                            ISRLLM~\cite{ISR-LLM:Zhou}{,}
                            ReAct~\cite{ReAct:Yao}{,}
                            Self-Refine~\cite{SelfRefine:AmanMadaan}{,}
                            Reflexion~\cite{Reflexion:NoahShinn}{,}
                            Plan{,} Verify and Switch~\cite{PlanVerifySwitch}{,}
                            ToT${^1}$~\cite{ToT:ShunyuYao}{,}
                            ToT${^2}$~\cite{ToT:JieyiLong}{,}
                            Tree-Planner~\cite{TreePlanner}{,}
                            RAP~\cite{RAP:Hao2023}{,}
                            LATS~\cite{LATS:Zhou}{,}
                            ToolChain*~\cite{toolchainstar}{,}
                            TPTU~\cite{TPTU}{,}
                            TPTUv2~\cite{TPTUv2}{,}
                            AgentInstruct~\cite{AgentInstructs:Crispino}{,}
                            ToRA~\cite{ToRA:Gou}{,}
                            AutoUI~\cite{AutoUI:Zhang}
                            , leaf, text width=55.0em
                        ]
                    ]
                    [
                        Distillation~(\S\ref{sec_frontiers_distill})
                        [
                            LMSI~\cite{SelfImprove:JiaxinHuang}{,}
                            STaR~\cite{STaR:Zelikman}{,}
                            \citet{TeachingSmallLM:LucieCharlotte}{,}
                            SCoTD~\cite{SCoTD:Li}{,}
                            SECToR~\cite{SECToR:Zhang}{,}
                            Distilling Step-by-Step~\cite{DistillingStep:Hsieh}{,}
                            SCOTT~\cite{SCOTT:Wang}{,}
                            DialCoT~\cite{DialCoT:Han}{,}
                            PlanningToken~\cite{PlanningToken:Wang}{,}
                            TailoredLearning~\cite{TailoredLearning}{,}
                            \citet{Self-ImproveThroughInteractiveDemonstrations}{,}
                            ImplicitCoT~\cite{ImplicitCoT}{,}
                            \citet{SpecializingSL:Fu}{,}
                            CoT Collection~\cite{CoTCollection:SeungoneKim}{,}
                            MoDE-CoTD~\citep{coling24-distill-moe-distill}{,}
                            PRR~\citep{coling24-distill-probe-retrieval}{,}
                            PaD~\citep{naacl24-distill-program}{,}
                            EoTD~\citep{wirr-distill-mixdistill-1}{,}
                            Mixed Distillation~\citep{wirr-distill-mixdistill-2}{,}
                            Math-Shepherd~\citep{wirr-distill-preference-deepseek}{,}
                            MCTS-IPL~\citep{wirr-distill-preference-mcts}
                            , leaf, text width=55.0em
                        ]
                    ]
                ]
                [
                    Future \\ Directions~(\S\ref{sec_future})
                    [
                        Multi-modal~(\S\ref{sec_future_multi_modal})
                        [
                            Multi-modalCoT~\cite{MMCOT:ZhuoshengZhang}{,}
                            GoT$^3$~\cite{GoT:YaoYao}{,}
                            ToMT~\cite{ToMT:PengboHu}{,}
                            Hypergraph-of-Thought~\cite{HoT:FanglongYao}{,}
                            T-SciQ~\cite{T-SciQ:LeiWang}{,}
                            SocraticQuestion~\citep{SocraticQuestion}{,}
                            MMReact~\citep{MMREACT}
                            , leaf, text width=55.0em
                        ]
                    ]
                    [
                        Faithfulness~(\S\ref{sec_future_faithfulness})
                        [
                            Rethinking and Retrievaling~\cite{RethinkingRetrieval:HangfengHe}{,}
                            Verify-and-Edit~\cite{VerifyEdit:RuochenZhao}{,}
                            CoK~\cite{CoK1:XingxuanLi}{,}
                            Verify-Edit~\citep{VerifyEdit:RuochenZhao}{,}
                            Chain-of-NLI~\citep{Chain-of-Natural-Language-Inference}{,}
                            \citet{DecomposeFaithful:AnshRadhakrishnan}{,}
                            \citet{MeasureFaithful:TameraLanham}{,}
                            \citet{SnowBall}
                            , leaf, text width=55.0em
                        ]
                    ]
                    [
                        CoT Theory~(\S\ref{sec_future_theory})
                        [
                            \citet{CoTEmpiricalStudy:BoshiWang}{,}
                            \citet{TextPattern}{,}
                            \citet{SemanticSymbolic:XiaojunTang}{,}
                            \citet{ExoressivePowerofTRFMCoT}{,}
                            \citet{AnalysisGradientCoT:SkylerWu}{,}
                            \citet{DissectingCoT:YingcongLi}{,}
                            \citet{TheoreticalPerspective:GuhaoFeng}{,}
                            \citet{why-llm-correct-cot}{,}
                            \citet{interpret-multi-step-reason}{,}
                            \citet{emnlp-best-paper}{,}
                            \citet{Emergent2:RylanSchaeffer}{,}
                            \citet{emergent-ability-survey}
                            , leaf, text width=55.0em
                        ]
                    ]
                ]
                [
                    Benchmarks~(\S\ref{sec_benchmarks})
                    [
                        Mathematical~(\S\ref{sec_benchmarks_math})
                        [
                            AddSub~\cite{AddSub:MohanmmadJavad}{,}
                            SingleEq~\cite{SingleEQ:Koncel-Kedziorski}{,}
                            MultiArith~\cite{MultiArith:RoySubhro}{,}
                            MAWPS~\cite{MAWPS:RikKoncel}{,}
                            AQUA-RAT~\cite{AQuA:WangLing}{,}
                            ASDiv~\cite{ASDiv:MiaoShenYun}{,}
                            SVAMP~\cite{SVAMP:ArkilPatel}{,}
                            GSM8K~\cite{Verifier:KarlCobbe}{,}
                            GSM-Hard~\cite{PAL:GapLuYu}{,}
                            MathQA~\cite{MathQA:AidaAmini}{,}
                            DROP~\cite{DROP:DuaDheeru}{,}
                            TheoremQA~\cite{TheoremQA:WenhuChen}{,}
                            TAT-QA~\cite{TATQA:FengbinZhu}{,}
                            FinQA~\cite{FinQA:ZhiyuChen}{,}
                            ConvFinQA~\cite{ConvFinQA:ZhiyuChen}{,}
                            MATH~\cite{MATH:DanHendrycks}{,}
                            NumGLUE~\cite{NumGLUE:SwaroopMishra}{,}
                            LILA~\cite{LILA:SwaroopMishra}{,}
                            Conic10K~\citep{conic10k}
                            , leaf, text width=55.0em
                        ]
                    ]
                    [
                        Commonsense~(\S\ref{sec_benchmarks_commonsense})
                        [
                            CSQA~\cite{CommonsenseQA:AlonTalmor}{,}
                            CSQA 2.0~\cite{CSQA2:AlonTalmor}{,}
                            ARC~\cite{ARC:SumithraBhakthavatsalam}{,}
                            OpenBookQA~\cite{OpenBookQA:MihaylovTodor}{,}
                            PIQA~\cite{PIQA:YonatanBisk}{,}
                            Event2Mind~\cite{Event2Mind:HannahRashkin}{,}
                            McTaco~\cite{McTaco:BenZhou}{,}
                            CosmosQA~\cite{CosmosQA:LifuHuang}{,}
                            ComValidation~\cite{ComV:WangCunxiang}{,}
                            ComExplanation~\cite{ComV:WangCunxiang}{,}
                            StrategyQA~\cite{StrategyQA}
                            , leaf, text width=55.0em
                        ]
                    ]
                    [
                        Symbolic~(\S\ref{sec_benchmarks_symbolic})
                        [
                            Last Letter Concat.~\cite{FewshotCoT:JasonWei}{,}
                            Coin Flip~\cite{FewshotCoT:JasonWei}{,}
                            Reverse List~\cite{FewshotCoT:JasonWei}{,}
                            BigBench~\cite{BigBench:AarohiSrivastava}{,}
                            BigBench-Hard~\cite{BigBenchHard:MiracSuzgun}
                            , leaf, text width=55.0em
                        ]
                    ]
                    [
                        Logical~(\S\ref{sec_benchmarks_logical})
                        [
                            ReClor~\cite{RClor:WeihaoYu}{,}
                            LogiQA~\cite{LogiQA:JianLiu}{,}
                            ProofWriter~\cite{ProofWriter:OyvindTafjord}{,}
                            FOLIO~\cite{FOLIO:SimengHan}{,}
                            PrOntoQA~\cite{PrOntoQA}{,}
                            LogiGLUE~\cite{logiglue}{,}
                            GLORE~\citep{glore}{,}
                            FALLACIES~\citep{naacl24-benchmark-logic-self-verification}
                            , leaf, text width=55.0em
                        ]
                    ]
                    [
                        Multi-modal~(\S\ref{sec_benchmarks_multi_modal})
                        [
                            Image-Text
                            [
                                ScienceQA~\cite{ScienceQA:PanLu}{,}
                                VCR~\cite{VCR:RowanZellers}{,}
                                VisualCOMET~\cite{VisualCOMET}{,}
                                PMR~\cite{PMR:QingxiuDong}{,}
                                CURE~\citep{multimodal-cot-dataset-cure}
                                , leaf, text width=44.5em
                            ]
                        ]
                        [
                            Video-Text
                            [
                                VLEP~\cite{VLEP:JieLei}{,}
                                CLEVRER~\cite{CLEVRER:KexinYi}{,}
                                STAR~\cite{STAR:BoWu}{,}
                                NExT-QA~\cite{NEXT-QA:JunbinXiao}{,}
                                Causal-VidQA~\cite{Causal-VidQA:JiantongLi}{,}
                                News-KVQA~\cite{NewsKVQA:PranayGupta}    
                                , leaf, text width=44.5em
                            ]
                        ]
                    ]
                ]
            ]
        \end{forest}
    }
    \caption{Taxonomy of Advanced Methods, Frontiers, Future Directions, and Benchmarks (Full Edition).}
    \label{fig:taxonomy_full}
\end{figure*}